\documentclass[11pt]{article}

\PassOptionsToPackage{table}{xcolor}
\usepackage[preprint]{acl}

\usepackage{times}
\usepackage{latexsym}
\usepackage{booktabs,tabularx}
\usepackage{float}
\usepackage{amsmath}
\usepackage{amssymb}
\usepackage{natbib}

\usepackage{booktabs}
\usepackage{tabularx}
\usepackage{xspace}

\definecolor{headerbg}{HTML}{F3DCC3} 
\definecolor{redbg}{HTML}{F6CACA}   
\definecolor{greenbg}{HTML}{DCEED6} 
\definecolor{diffbg}{HTML}{E6DDF2}   
\definecolor{goldvar}{HTML}{B8860B}
\definecolor{predvar}{HTML}{555555}
\definecolor{questionvar}{HTML}{111111}
\definecolor{functionvar}{HTML}{003B73} 
\newcommand{\question}{\ensuremath{\textcolor{questionvar}{\mathsf{q}}}}
\newcommand{\gold}{\ensuremath{\textcolor{goldvar}{\mathsf{g}}}}
\newcommand{\pred}{\ensuremath{\textcolor{predvar}{\mathsf{p}}}}
\newcommand{\labelclass}{\ensuremath{\textcolor{questionvar}{\mathsf{c}}}}

\newcommand{\goldstatement}{\mathbf{s}_{\gold}}
\newcommand{\predstatement}{\mathbf{s}_{\pred}}

\newcommand{\astatement}{\mathbf{s}_{a}}
\newcommand{\bstatement}{\mathbf{s}_{b}}
\newcommand{\statement}[2]{%
  \ensuremath{\textcolor{questionvar}{\mathbf{s}}\!\left(#1,#2\right)}%
}

\newcommand{\jscoreempty}{%
  \ensuremath{\textcolor{questionvar}{\mathbf{D}}}%
}

\newcommand{\jscore}[2]{%
  \ensuremath{\textcolor{questionvar}{\mathbf{D}}\!\left(#1 \rightarrow #2\right)}%
}

\newcommand{\capfunction}[2]{%
  \ensuremath{\textcolor{functionvar}{\mathbf{CAP}}\!\left(#1,#2\right)}%
}
\newcommand{\metricempty}{%
  \ensuremath{\textcolor{functionvar}{\mathbf{m}}}%
}
\newcommand{\metric}[2]{%
  \ensuremath{\metricempty\!\left(#1,#2\right)}%
}
\newcommand{\rankfunc}[1]{%
  \ensuremath{\textcolor{questionvar}{\mathbf{r}}\!\left(#1\right)}%
}

\newcommand{\pairacc}[1]{%
  \ensuremath{\textcolor{questionvar}{\mathbf{PairAcc}}\!\left(#1\right)}%
}
\usepackage{dsfont}
\newcommand{\indicator}[1]{%
  \ensuremath{\mathds{1}_{\{#1\}}}%
}

\newcommand{\meanscore}[2]{%
  \ensuremath{\textcolor{questionvar}{\boldsymbol{\mu}}_{#1}\!\left(#2\right)}%
}

\newcommand{\violationrate}[1]{%
  \ensuremath{\textcolor{questionvar}{\mathbf{V}}\!\left(#1\right)}%
}

\usepackage[T1]{fontenc}

\usepackage[utf8]{inputenc}

\usepackage[nopatch=footnote]{microtype}

\usepackage{inconsolata}

\usepackage{graphicx}
\usepackage[disable]{todonotes}

\usepackage{setspace}

\usepackage{tabularray} 
\usepackage{hyperref}   

\newcommand{\openqa}{Open QA}

\usepackage[capitalize,noabbrev]{cleveref}
\crefname{section}{Sec.}{Secs.}
\Crefname{section}{Sec.}{Secs.}

\crefname{table}{Tab.}{}
\crefname{figure}{Fig.}{Figs.}
\crefname{algorithm}{Alg.}{}
\crefname{equation}{Eq.}{}
\crefname{appendix}{App.}{Apps.}
\Crefname{appendix}{App.}{Apps.}
\crefname{thm}{Theorem}{}
\crefname{prop}{Proposition}{}
\crefname{defin}{Definition}{}
\crefname{cor}{Corollary}{}
\crefname{observation}{Observation}{}
\crefname{assumption}{Assumption}{}
\crefname{sublemma}{SubLemma}{SubLemmas}
\crefformat{footnote}{#2\footnotemark[#1]#3}

\AddToHook{cmd/appendix/before}{%
    \crefalias{section}{appendix}%
    \crefalias{subsection}{appendix}%
    \crefalias{subsubsection}{appendix}%
}
\title{How Correct Is Your Answer?\\ A Semantic Correctness Framework for Open QA Evaluation}

\author{
\textbf{Elitsa Yotkova$^1$\thanks{Equal contribution.},
Violeta Kastreva$^{1,2}$\footnotemark[1],
Petar Velkov$^{1,3}$,
Hristo Boyanov$^1$,} \\
\textbf{Dimitar Dimitrov$^1$,
Preslav Nakov$^4$,
Ivan Koychev$^1$} \\
$^1$Sofia University ``St. Kliment Ohridski'',
$^2$ETH Zürich, 
$^3$University of Zurich, \\
$^4$Mohamed bin Zayed University of Artificial Intelligence \\
\texttt{eyotkova@g.fmi.uni-sofia.bg, vkastreva@ethz.ch, petar.velkov@uzh.ch} \\
\texttt{hbojanov@uni-sofia.bg, \{ilijanovd, koychev\}@fmi.uni-sofia.bg} \\
\texttt{preslav.nakov@mbzuai.ac.ae}
}

\begin{document}
\maketitle

\begin{abstract}
Reliable evaluation of open-ended question answering remains a bottleneck for measuring answer correctness of modern LLMs. Unlike multiple-choice tasks, free-form answers may be correct in many surface forms and may fail in qualitatively different ways, including incompleteness, contradiction, overgeneration, and endorsement of false premises. Existing judgment-based and similarity-based metrics often collapse these distinctions.
We address this gap with three reusable contributions. First, we introduce a semantic correctness taxonomy that assigns open-ended answers to eight ordered classes, separating verbose-but-correct answers from those contaminated by hallucinated content. Second, we release CAP-Correctness, an 8.8k-example benchmark spanning widely used QA datasets, and CAP-Statements, an 11k-example dataset for converting question-answer pairs into declarative statements for natural language inference (NLI) training and statement-based evaluation. Third, we introduce CAP (Context-Aware Precision), a reference-based metric that scores question-conditioned statements using bidirectional NLI.
Under a monotonicity protocol testing whether metrics respect the taxonomy’s intended ordering, CAP outperforms established baselines.\footnote{Code and data are available at \url{https://github.com/ElitsaY/EMNLP-2026-How-Correct-Is-Your-Answer-}.} 

\end{abstract}

\section{Introduction}


Open-ended question answering (Open QA) is a long-established task requiring systems to generate free-form answers to questions across diverse domains \citep{fan-etal-2019-eli5}. Unlike multiple-choice question answering (MCQA), where answers are restricted to predefined options, Open QA permits unconstrained free-form responses, making automatic correctness evaluation substantially more challenging. In Open QA evaluation, free-form model outputs often lack explicit ground-truth correctness labels, making it difficult to determine whether a prediction is correct. Even when a ground-truth answer is available--which we refer to as the gold answer--evaluation remains challenging. 
\begin{figure}[t]
    \centering
    \includegraphics[width=0.5\textwidth]{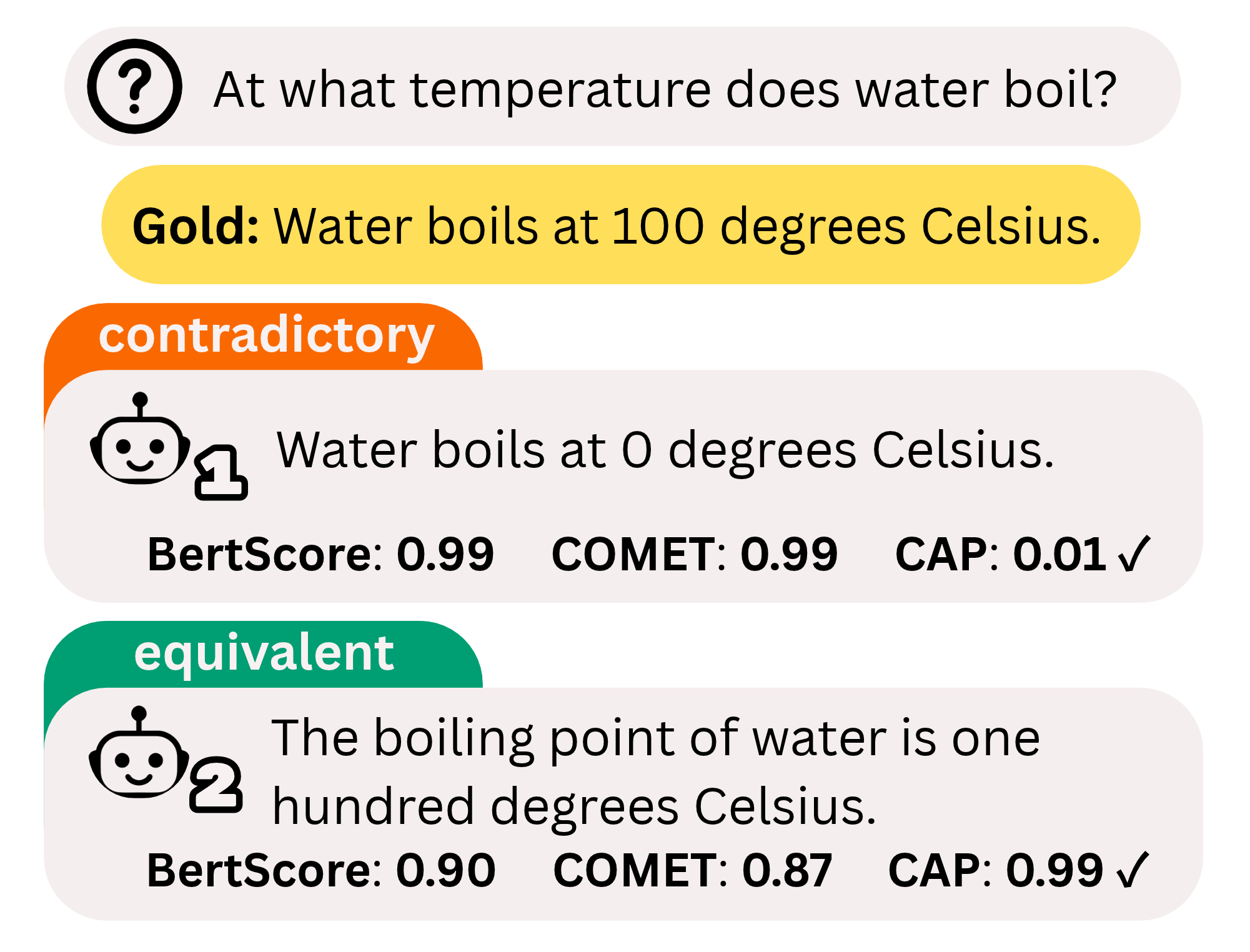}
    \caption{Comparison of QA evaluations, including our proposed CAP.}
    \label{figure:answer-example}
\end{figure}



A prediction can be correct while sharing few tokens with the reference \citep{bulian-etal-2022-tomayto}, while a lexically similar prediction may omit required information, add unsupported claims, or contradict the gold answer. \cref{figure:answer-example} illustrates this limitation: 
similarity-based metrics may reward contradictory answers while penalizing paraphrased correct ones. Evaluating \openqa{}, therefore, requires modeling correctness beyond textual similarity. 


These shortcomings have also appeared in shared-task evaluations. In ClinIQLink 2025 \citep{colelough-etal-2025-overview}, BLEU \citep{papineni-etal-2002-bleu} fails to recognize correct paraphrased answers, motivating a task-specific semantic scoring scheme. Similarly, AraHealthQA 2025 \citep{alhuzali-etal-2025-arahealthqa} uses BERTScore \citep{zhang2020bertscore} for open-ended answers, even though the organizers note that automatic metrics do not fully capture answer appropriateness or trustworthiness.
 Together, these tasks illustrate the need for a {reproducible Open QA evaluation framework}.

QA-specific evaluation protocols often reduce generated answers to exact-match scores or binary accept/reject judgments. This is too coarse for \openqa{}, where candidate answers may be incomplete, overinclusive, unsupported, contradictory, or based on a false premise \citep{kamalloo-etal-2023-evaluating, adlakha-etal-2024-evaluating, yao-barbosa-2024-accurate}. Such binary labels obscure the type of error, especially for LLM outputs that combine correct content with additional explanations.


A natural alternative
to these protocols is to use LLMs themselves as judges, since they can provide more flexible semantic assessment. However, LLM-as-a-judge methods are costly to run at scale, sensitive to prompting choices, less reproducible, and affected by known biases such as position and verbosity effects \citep{zheng2023judging, shi2024judging}.

To address these limitations,
we propose a fine-grained semantic correctness taxonomy for reference-based \openqa{} evaluation. Rather than reducing generated answers to binary correct/incorrect labels, we define a taxonomy consisting of eight classes that distinguish semantic relations among candidate answers, reference answers, and questions. This enables more diagnostic evaluation by identifying not only whether an answer is correct, but how it relates to the expected answer. To further capture gradations in semantic correctness, we impose an ordering over these classes that reflects their expected relative quality, enabling us to evaluate whether metrics behave monotonically as answers become less correct.

Building on this taxonomy, we introduce a semantic correctness framework for \openqa{} evaluation that uses our proposed reference-based metric Context-Aware Precision (CAP). Given a question, a gold answer, and a predicted answer, CAP reformulates each question–answer pair as a question-conditioned declarative statement. Then it compares the resulting statements using bidirectional natural language inference (NLI). NLI offers a lightweight, reproducible alternative for studying the limitations of pretrained evaluators. CAP produces a continuous semantic correctness score in $[0,1]$ that can be analyzed against the taxonomy's ordering.

We evaluate CAP in a controlled English short-form QA setting using CAP-Correctness, an 8.8k-example semantic correctness dataset derived from OpenBookQA, ARC, and MMLU \citep{mihaylov-etal-2018-suit, clark2018think, hendrycks2021measuring}. The resource covers diverse domains, question formats, and answer relations. Since CAP applies NLI to declarative statements rather than raw question–answer pairs, we also construct CAP-Statements, an 11k-example dataset for question-conditioned QA-to-statement reformulation, and manually evaluate the reliability of the generated statements.


Using these resources, we compare CAP against commonly used 
lexical-overlap, embedding-based, and learned semantic metrics by testing whether their scores preserve the intended ordering of correctness categories. Across comparisons, CAP achieves stronger semantic ranking alignment, higher pairwise ordering accuracy, and fewer monotonicity violations
, indicating that it better reflects the taxonomy’s intended correctness ordering than similarity-based alternatives.

Our contributions are as follows:





\begin{itemize}
\item We propose an eight-class semantic correctness taxonomy for \openqa{} evaluation that distinguishes qualitatively different answer behaviors, including partial answers, valid elaboration, hallucination-contaminated answers, and contradictions. We instantiate the taxonomy in CAP-Correctness, an 8.8k-example controlled semantic correctness benchmark. 

    \item We introduce CAP, a reference-based semantic correctness metric based on bidirectional NLI over question-conditioned declarative statements. To support its statement-reformulation step, we construct CAP-Statements, an 11k-example dataset for training and evaluating QA-to-statement generation.

    \item We establish a monotonicity-based diagnostic protocol for evaluating whether metric scores respect the intended ordering of semantic correctness classes, revealing systematic failure modes across existing evaluators.

\end{itemize}



\begin{figure*}[!th]
    \centering
    \includegraphics[width=1.0\textwidth]{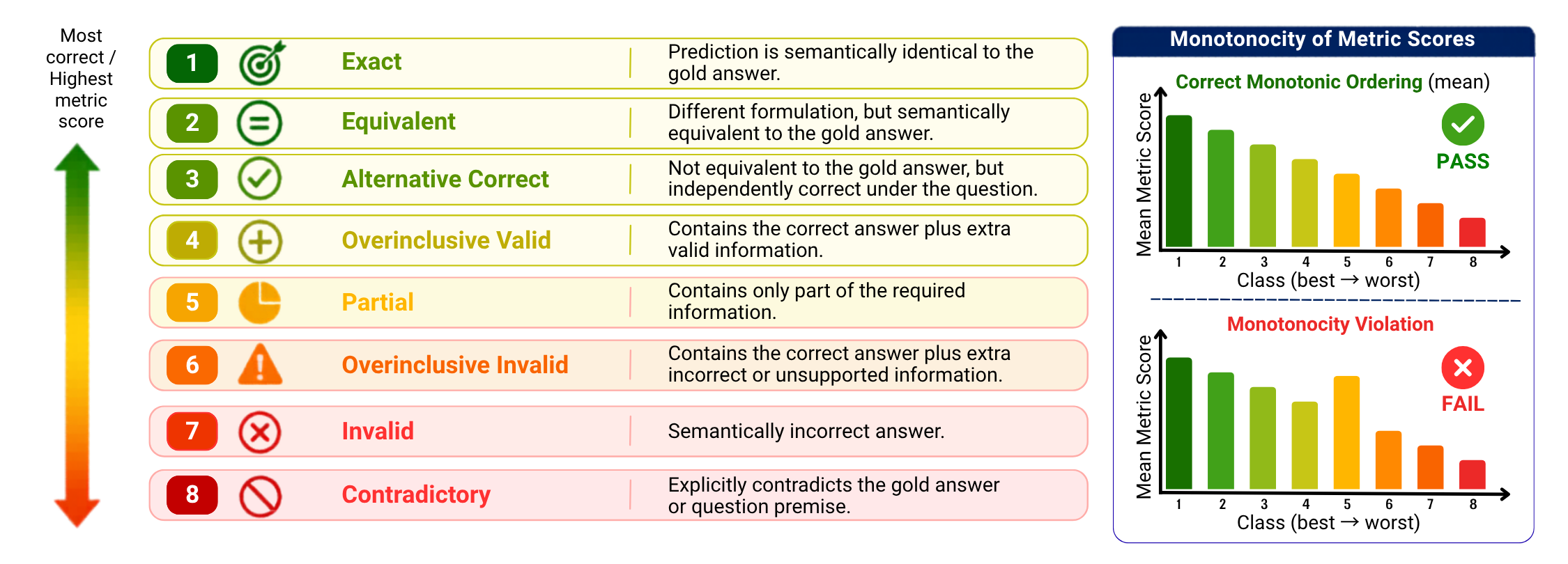}    \caption{Correctness taxonomy for \openqa{} and monotonicity criterion.
We define eight answer classes, ordered from most to least correct. A metric is considered higher quality if its mean score decreases monotonically across this ordering.}
    \label{fig:pipeline}
\end{figure*}

\section{Related Work}



\subsection{NLI-based Evaluation}

Natural language inference has been used to evaluate generated answers beyond
lexical similarity. \citet{chen-etal-2021-nli-models} formulate QA verification
as an entailment problem over questions, evidence, and predicted answers,
treating correctness as an entailed-or-not decision. Building on this,
\citet{honovich-etal-2022-true} benchmark NLI-based and QA-based metrics across
eleven factual consistency datasets, finding large-scale NLI to be among the
strongest evaluation approaches. \citet{laban-etal-2022-summac} further show
that decomposing documents into sentence-level units before applying NLI yields
more reliable inconsistency detection, motivating fine-grained statement-level
evaluation. \citet{zha-etal-2023-alignscore} unify NLI, QA, and
fact-verification signals into a single alignment function that achieves
GPT-4-level factual consistency scoring. \citet{chen-eger-2023-menli}
demonstrate that NLI-based metrics are more robust than embedding-similarity
metrics under meaning-changing perturbations, providing evidence that
directional inference captures semantic content more faithfully than symmetric
similarity.

Building on this line of work, CAP uses bidirectional NLI within a taxonomy-driven diagnostic framework that distinguishes equivalence, incompleteness, and overgeneration. We further introduce a monotonicity benchmark that tests whether a metric's scores respect the ordering induced by our correctness taxonomy, enabling systematic comparison across evaluation methods. 

\subsection{Correctness Taxonomies for \openqa{}}
Early \openqa{} evaluation relied on exact match or lexical overlap, which
systematically misclassifies paraphrased correct answers as wrong and rewards
surface-similar incorrect ones \citep{papineni-etal-2002-bleu, lin-2004-rouge}.
\citet{kamalloo-etal-2023-evaluating} show that these metrics underestimate
true QA accuracy by over 50\% on modern LLM outputs and that existing learned
evaluators also fail on free-form answers. \citet{adlakha-etal-2024-evaluating}
further show that correctness and faithfulness are distinct aspects of model
outputs that coarse evaluation schemes may conflate.

More recent work has moved toward graded correctness and answer equivalence.
\citet{bulian-etal-2022-tomayto} propose entailment-based answer equivalence criteria
that accept any answer containing at least all required content without
misleading additions. 
\citet{yona-etal-2024-granola} focus on variation among \emph{valid} answers at different levels of specificity, evaluating predictions for accuracy and informativeness against explicitly constructed multi-granularity reference answers. 
\citet{yao-barbosa-2024-accurate} instead characterize answers by their entailment relation to the gold answer, using NLI signals within an LLM-based evaluator to assign partial or bonus credit according to the inference gap between answers.

For evaluating verbose LLM-generated answers, however, two important gaps remain. 
First, none explicitly separates a correct answer
accompanied by valid elaboration from a correct answer contaminated by
hallucinated content--a distinction central to evaluating modern LLM outputs. 
\citet{min-etal-2023-factscore} address the related problem of factual precision
in long-form text by decomposing outputs into atomic facts and verifying each
independently, but this approach operates at sub-sentence granularity and does
not yield an answer-level correctness class. Second,
existing partial-credit
mechanisms either rely on LLM-generated intermediate reasoning to estimate
inference difficulty \citep{yao-barbosa-2024-accurate}, or on token-level
overlap, tying the score to surface form rather than to direct semantic
relations between expected and predicted answers.

CAP helps address these gaps.
We extend the correctness taxonomy to eight
classes by adding \texttt{overinclusive-valid} and
\texttt{overinclusive-invalid} categories that capture the verbose behavior
characteristic of modern LLM outputs, making the evaluation diagnostic: a
correct but verbose answer should not be penalized the same way as a correct
answer contaminated by hallucinated content. Rather than eliciting post-hoc
reasoning for partial credit, CAP derives a continuous score from bidirectional
NLI probabilities over question-conditioned declarative statements, tying the
score directly to semantic compatibility and completeness.

\section{Proposed Taxonomy for OpenQA}\label{sec:taxonomy}
We define semantic correctness as the relation between a predicted answer and a gold answer under the same question context. To evaluate an open-ended answer, exact matching is only the simplest case: a prediction may repeat the gold answer verbatim, giving an \texttt{exact} answer, or express the same meaning with different wording, giving an \texttt{equivalent} answer. However, \openqa{} often allows answers that are correct and differ greatly from the reference. For example, for a question such as \textit{Name three cities in Europe}, many completely different sets of cities may still be correct; we treat these as \texttt{alternative-correct} answers \cref{fig:pipeline}. 



Model responses can fail or deviate from the target answer in distinct ways. A \texttt{partial} answer provides only some of the required information, such as naming two cities when three are requested. An \texttt{overinclusive-valid} answer includes extra information beyond the request, but the added content is accurate and relevant—for example, briefly describing the cities’ locations.
If the response contains the correct answer but also introduces an unsupported or false claim, such as listing a non-European city, it becomes \texttt{overinclusive-invalid}. Other predictions may directly contradict the gold answer or the question premise, giving a \texttt{contradictory} answer, while remaining mistakes that do not fit these cases are labeled \texttt{invalid} (\cref{fig:pipeline}). These distinctions also define an expected ordering of answer quality, shown in \cref{eq:correctness-order}. Fully correct answers should receive higher semantic correctness scores than incomplete answers, and incomplete answers should generally score above answers that introduce false information or contradict the question. We formalize this expectation as a \textbf{monotonicity property}, which is used throughout our evaluation. 
\begin{equation}
\label{eq:correctness-order}
\begin{aligned}
\texttt{exact}
&\approx
\texttt{equivalent}
\approx
\texttt{alternative-correct}
\\
&\geq
\texttt{overinclusive\_valid}
>
\texttt{partial}
\\
&>
\texttt{overinclusive\_invalid}
\\
&>
\texttt{invalid}
\geq
\texttt{contradictory}.
\end{aligned}
\end{equation}
We use this ordering as the benchmark's monotonicity target; neighboring relations such as \texttt{partial} and \texttt{overinclusive-valid} may be adapted to application-specific preferences.


\section{CAP-QA Framework}\label{sec:cap-framework}

An NLI model maps an ordered statement pair $(\astatement, \bstatement)$ to a distribution over three labels--\emph{entailment}, \emph{neutral}, and \emph{contradiction}. We write $\astatement \rightarrow \bstatement$ for the directed inference from premise $\astatement$ to hypothesis $\bstatement$, and obtain these probabilities from a pretrained NLI classifier (see \cref{app:implementation-details}).
\paragraph{CAP Design.}
Given a question $\question$, a gold answer $\gold$, and a predicted answer $\pred$, we first generate corresponding declarative statements $\statement{\question}{\gold}$  and $\statement{\question}{\pred}$. CAP then measures the semantic relationship between the generated statements using bidirectional entailment scoring.
Let $\goldstatement := \statement{\question}{\gold}$ and $\predstatement := \statement{\question}{\pred}$. For these statements, we define a directional score:
\begin{align}
    &\jscore{\goldstatement}{\predstatement} := \\ \nonumber
&P_{\text{entailment}}(\goldstatement \rightarrow \predstatement)
+
\lambda P_{\text{neutral}}(\goldstatement \rightarrow \predstatement),
\end{align}

where \(\lambda \in [0,1]\) controls the contribution of the neutral class. CAP is then defined as:
\begin{align}
&\capfunction{\goldstatement}{\predstatement}\ := \\ \nonumber
&\alpha \jscore{\goldstatement}{\predstatement}
+
(1-\alpha) \jscore{\predstatement}{\goldstatement},
\end{align}

where \(\alpha \in [0,1]\) balances semantic compatibility and semantic completeness. 



The bidirectional formulation allows CAP to distinguish between
semantically equivalent and semantically incomplete answers by
incorporating entailment in both directions between the gold and
predicted statements. In our experiments, we use  $\alpha = 0.85$ and $\lambda = 0.30$. These values were selected via a sweep on a held-out subset of CAP-Correctness;
moderate neutral weighting combined with asymmetric scoring jointly
maximizes semantic ranking. The full sweep is reported in
\cref{app:cap-sweep}.



\paragraph{Statement Generation.}
To reformulate question–answer pairs into declarative statements for NLI evaluation, we fine-tune an mT5-based seq-to-seq model \citep{xue-etal-2021-mt5} on our statement generation dataset described in \cref{sec:datasets}. Fine-tuning details are provided in \cref{sec:datasets} and \cref{app:experiment-details}. We use the mT5 model to provide a cheaper, standardized, and reproducible reformulation component for CAP. On the held-out test set, the model achieves strong surface-form agreement with the references: 96.64 BLEU, 98.08 ROUGE-L, and 76.7\% Exact Match. Manual inspection confirms that most non-exact outputs preserve the intended meaning, making the generated statements suitable for downstream NLI evaluation. 




\paragraph{Boundedness.} CAP is a convex combination of two
directional scores $\jscoreempty \in[0,1]$ and therefore yields a
continuous score $\capfunction{\astatement}{\bstatement}\in[0,1]$. Full derivation is
given in \cref{app:cap-bounded}.


\section{Dataset Construction}\label{sec:datasets}
We instantiate our evaluation framework with two complementary datasets: \textbf{(1) CAP-Correctness}, a semantic evaluation annotated using our proposed correctness label set, and \textbf{(2) CAP-Statements}, a dataset for reformulating question--answer pairs into declarative statements for NLI-based evaluation. Details on dataset acquisition, annotation, and statistics are provided in \cref{app:dataset}.

\begin{table}[H]
\centering
\small
\setlength{\tabcolsep}{4pt}
\renewcommand{\arraystretch}{1.05}
\begin{tabular}{lccc}
\rowcolor{diffbg}
\toprule
\textbf{Dataset} & \textbf{Train} & \textbf{Val.} & \textbf{Test} \\
\midrule
CAP-Correctness & - & 1000 & 7827 \\
CAP-Statements & 8800 & 1100 & 1100 \\
\bottomrule
\end{tabular}
\caption{Dataset statistics.}
\label{tab:datasetsshort}
\end{table}




\textbf{CAP-Correctness} contains 8{,}827 \emph{[question, gold answer, candidate answer, correctness label]} examples from OpenBookQA, AI2 ARC, and MMLU, spanning elementary through undergraduate-level questions. We clean the collected data by removing questions such as "Which of the following". Moreover, we remove gold answers in a form of "all of above", "none of above" which are MCQ specific. Candidate answers and labels are produced by an LLM-assisted pipeline, with a 1638-example subset (18.56\%) re-labeled by human annotators against the same taxonomy. Human--LLM agreement is substantial (\cref{tab:agreement}: Cohen's $\kappa = 0.779$, quadratic-weighted $\kappa = 0.879$), confirming that the synthetic labels reliably track human judgements
of semantic correctness. Full annotation protocol, along with statistics is provided in \cref{app:cap-correctness}.

\textbf{CAP-Statements} is a collection of 11{,}000
\emph{[question, answer, statement]} triples, where the statement
preserves the semantic content of its corresponding question--answer
pair. The dataset spans four question types
(\cref{tab:question-type-labels}); among these, long-form items with
multi-sentence context are the hardest reformulation regime, since
several sentences must be compressed into a single declarative claim.
This makes CAP-Statements a non-trivial benchmark for statement
generation as well as a natural supervision source for
QA-to-statement reformulation \citep{demszky2018transformingqa}.
Details on construction and human validation are provided in
\cref{app:cap-statements}.





\section{Experiment Design}\label{sec:experiments}


We evaluate whether CAP and existing automatic metrics preserve the semantic-correctness ordering induced by our taxonomy (\cref{eq:correctness-order}). The comparison covers widely used metrics from three families:  \textbf{Lexical overlap:} BLEU,
    ROUGE-L, METEOR
    \citep{banerjee-lavie-2005-meteor}. 
     \textbf{Contextual embedding:} BERTScore (F1).
\textbf{Learned semantic regression:} COMET    \citep{rei-etal-2020-comet}.

For each metric we report the exact model checkpoint and version
used in \cref{app:experiment-details}.

\subsection{Research Questions}

Our experiments are designed around three research questions:

\begin{itemize}
    \item \textbf{RQ1 (Monotonicity)} For every class pair
    $(c_i, c_j)$ with $c_i \succ c_j$ in our taxonomy, does a mean
    metric score on $c_i$ exceed the mean on $c_j$?
    \item \textbf{RQ2 (Local separability)} Do metrics distinguish
    \emph{neighboring} taxonomy classes that are especially
    challenging for surface- and embedding-based scorers?
    
    \item \textbf{RQ3 (Generalization to LLM outputs)} Does CAP
    preserve these properties on free-form answers generated by
    state-of-the-art LLMs, rather than only on the semi-synthetic
    answer variants in CAP-Correctness?
\end{itemize}




\subsection{Evaluation Measures}\label{sec:eval-measures}

For every gold--prediction--label triple
$(\gold, \pred, \labelclass) \in$ CAP-Correctness, a metric $\metricempty$
produces a score $\metric{\gold}{\pred} \in [0,1]$. We evaluate whether these
scores respect the semantic correctness ordering in \cref{eq:correctness-order}
using four complementary measures. \textbf{Rank correlation} measures global agreement between metric scores and taxonomy ranks using Spearman's $\rho$ and Kendall's $\tau$. \textbf{Pairwise ranking accuracy} reports the fraction of class-ordered answer pairs for which the metric assigns a higher score to the more correct answer (random baseline $0.5$). \textbf{Monotonicity violations} count class pairs whose mean metric scores invert the expected taxonomy order. \textbf{Hard neighboring-pair accuracy} restricts pairwise accuracy to adjacent, locally difficult class contrasts in \cref{eq:correctness-order}. Full definitions are provided in \cref{app:eval-measures}.

\subsection{Implementation Details}\label{sec:implementation}

CAP scores are computed with \texttt{cross-encoder/ nli-deberta-v3-large} over declarative statements produced by an mT5-small generator fine-tuned on the CAP-Statements training split. We use the 1,000-example CAP-Correctness validation split for CAP hyperparameter and NLI-backbone selection. All headline comparisons in §7.1–§7.3 are then reported on the untouched 7,827-example test split. Full checkpoint pins, baseline versions, and bootstrap details are in \cref{app:implementation-details}.

\section{Results} \label{sec:results}

\subsection{CAP against Established Metrics}\label{sec:main-results}
\cref{tab:main-ranking-results} reports the headline comparison. CAP operates in a markedly different regime from the baselines. Lexical metrics and BERTScore remain close to the $0.5$ pairwise-accuracy baseline and show weak rank correlations, suggesting that surface-form and embedding similarity provide little signal for our taxonomy’s ordering. COMET performs better than this near-random band, but still recovers only part of the ordering, consistent with its tendency to merge answer relations that our taxonomy distinguishes. CAP achieves the strongest rank correlation by a wide margin and raises pairwise accuracy into a clearly informative range (\cref{app:confidence}).
\begin{table}[!tbh]
\centering
\setlength{\tabcolsep}{4pt}
\renewcommand{\arraystretch}{1.05}
\footnotesize
\begin{tabular}{lccc}
\toprule
\rowcolor{diffbg}
\textbf{Metric} & \textbf{Spearman $\rho$} & \textbf{Kendall $\tau$} & \textbf{Pairwise Acc.} \\

\midrule
BLEU      & 10.95  & 8.31  & 51.57  \\
ROUGE-L   & 16.67  & 13.00  & 55.35  \\
METEOR    & 14.70  & 11.45  & 53.71  \\
\rowcolor{redbg}
BERTScore & 16.10  & 10.90  & 56.18  \\
\rowcolor{redbg}
COMET     & 26.88  & 19.57  &  61.10  \\
\midrule
\rowcolor{greenbg}
\textbf{CAP} & \textbf{60.37} & \textbf{48.83} & \textbf{77.70} \\
\bottomrule
\end{tabular}
\caption{Correlation and pairwise ranking accuracy between metric scores and semantic correctness ordering.}
\label{tab:main-ranking-results}
\end{table}

The per-pair separability profile in \cref{tab:pairwise-auc} places this gain on a difficulty axis: CAP is near-perfect on distant class pairs and degrades smoothly as the pair becomes more local. 

\begin{table}[!tbh]
\centering
\setlength{\tabcolsep}{5pt}
\renewcommand{\arraystretch}{1.05}
\footnotesize
\begin{tabular}{lc}
\toprule
\rowcolor{diffbg}
\textbf{Comparison} & \textbf{CAP AUC} \\
\midrule
Exact $>$ Invalid & 98.24 \\
Partial $>$ Invalid & 94.39 \\
Equivalent $>$ Invalid & 92.63 \\
Equivalent $>$ OV & 80.65 \\
Partial $>$ OI & 89.03 \\
OV $>$ OI & 69.48 \\
\bottomrule
\end{tabular}
\caption{Pairwise semantic separability of CAP across semantic correctness categories.}
\label{tab:pairwise-auc}
\end{table}

The next two subsections decompose this picture at the class-mean level (\cref{sec:monotonicity}) and at the per-example level on the hardest neighbors (\cref{sec:hard-pairs}).

\subsection{Monotonicity Analysis}\label{sec:monotonicity}

\begin{figure}[!th]
    \centering
    \includegraphics[width=0.45\textwidth]{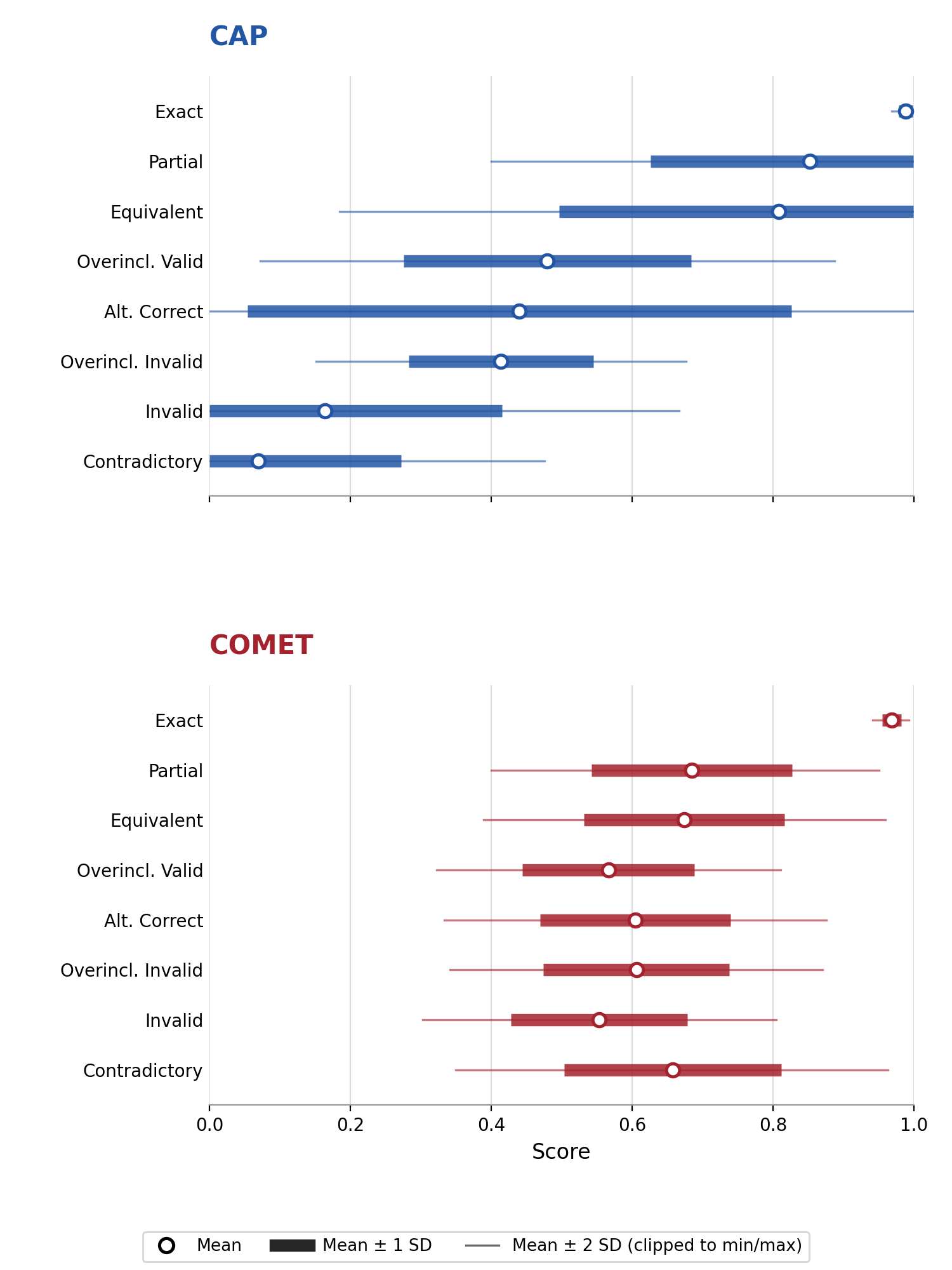}
    \caption{Geometric representation of CAP vs COMET.}
    \label{fig:comet_cap_distribution}
\end{figure}

At the class-mean level, CAP also largely preserves the taxonomy's
\emph{ordering}, not just its global rank correlation. The
per-pair separability profile in \cref{tab:pairwise-auc} shows
that CAP is near-perfect on distant class pairs and degrades
smoothly toward the local ones, and \cref{fig:comet_cap_distribution}
makes the geometry visible: CAP's per-class distributions form
largely distinct bands on the score axis, while COMET's collapse
into a narrow overlapping region. Across the 25 strictly ordered
class pairs, lexical metrics and BERTScore invert roughly half,
COMET inverts $9/25$, and CAP reduces this to $4/25$, with the
remaining inversions concentrated on \texttt{alternative-correct}
and the \texttt{partial} / \texttt{overinclusive-valid} pair.
Full counts, per-class means, and mechanism analysis are in
\cref{app:monotonicity}. At the same time, these residual failures indicate that improved ordering should not be conflated with reliable correctness assessment: none of the evaluated metrics, including CAP, consistently resolves all semantic distinctions in the taxonomy.


 



\subsection{Hard Neighboring-Pair Evaluation}\label{sec:hard-pairs}

While monotonicity captures the ordering at the class-mean level, pairwise accuracy on neighboring class pairs (\cref{tab:hard-pairwise-ranking}) tests whether the ordering holds at the per-example level on the hardest contrasts. Most baselines collapse to or below chance here, confirming that their above-baseline global behavior in \cref{tab:main-ranking-results} is carried by easy pairs at the extremes of the ordering---\texttt{equivalent} vs.\ \texttt{invalid}, \texttt{exact} vs.\ \texttt{contradictory}---and not by the genuinely hard distinctions in the middle of the taxonomy.
\begin{table}[tbh]
\centering
\setlength{\tabcolsep}{3pt}
\renewcommand{\arraystretch}{1.05}
\footnotesize
\begin{tabular}{lcccc}
\toprule
\rowcolor{diffbg}
\textbf{Metric} & \textbf{Eq $>$ Part} & \textbf{OV $>$ Part} & \textbf{OV $>$ OI} & \textbf{Alt $>$ Inv} \\
\midrule
BLEU      & 36.47 & 51.42 & 34.80 & 36.78 \\
ROUGE-L   & 23.21 & 27.37 & 27.74 & 35.74 \\
METEOR    & 41.31 & \textbf{70.66} & 36.38 & 39.86 \\
BERTScore & 51.86 & 32.02 & 30.88 & 57.21 \\
COMET     & 47.36 & 26.88 & 41.61 & 61.01 \\
\midrule
\rowcolor{greenbg}
\textbf{CAP} & \textbf{59.73} & 16.34 & \textbf{69.48} & \textbf{73.20} \\
\bottomrule
\end{tabular}
\caption{Pairwise ranking accuracy on hard neighboring semantic distinctions. OV = overinclusive-valid, OI = overinclusive-invalid, Alt = alternative-correct, Inv = invalid.}
\label{tab:hard-pairwise-ranking}
\end{table}

The \texttt{overinclusive-valid} vs. \texttt{partial} comparison is the
most informative failure case. All metrics except METEOR fall below chance
on this pair, with CAP showing the largest reversal. This is a direct
consequence of CAP's asymmetric bidirectional formulation
(\cref{sec:cap-framework}): partial answers retain high
$\gold \!\rightarrow\! \pred$ entailment along the heavily-weighted
direction ($\alpha = 0.85$), while overinclusive-valid answers reverse
this asymmetry. \cref{tab:per-class-label-source} and
\cref{fig:comet_cap_distribution} confirm the resulting inversion across
both label sources. 
This pair remains the clearest diagnostic of CAP's design tradeoff: the same asymmetry that helps CAP elsewhere makes it near-inverted on this axis. 

Per-unit entailment over atomic semantic decompositions could expose this asymmetry more directly than whole-statement NLI; we discuss this further in \cref{sec:conclusion}.

\subsection{Evaluation against LLM outputs}\label{sec:llm-eval}


To assess external validity, we test whether our eight-class taxonomy captures how current LLMs answer open-ended questions and whether CAP’s class-mean ordering holds for model-generated answers. We collect zero-shot responses from GPT-4o, Gemini 2.0 Flash, and Qwen3-8B-Instruct on a random sample of 1,000 CAP-Correctness questions. Human annotators then label each response according to the taxonomy (see \cref{app:annotation}).

\begin{table}[!tbh]
\centering
\scriptsize

\begin{tabular}{lccc}

\toprule

\rowcolor{diffbg}

\textbf{Class} & \textbf{GPT-4o} & \textbf{Gemini Flash} & \textbf{Qwen 3} \\

\midrule

exact                 & 97.27 & 91.20 & 98.87 \\

equivalent            & 78.81 & 70.66 & 80.18 \\

alternative-correct   & 31.59 & 26.77 & 29.65 \\

overinclusive-valid   & 40.32 & 44.08 & 47.47 \\

partial               & 35.64 & 48.87 & 46.87 \\

overinclusive-invalid & 36.58 & 38.82 & 28.38 \\

invalid               & 25.70 & 38.75 & 23.26 \\

contradictory         & 18.18 & 28.16 & 4.53 \\

\bottomrule

\end{tabular}

\caption{Mean CAP score on human-labeled LLM-generated answers, grouped by assigned correctness class.} 

\label{tab:llm-semantic-distribution}

\end{table}

The resulting annotations support both claims. First, LLM responses populate all eight taxonomy classes (\cref{tab:llm-semantic-counts}). This suggests that the proposed categories capture real model behavior, not only CAP-Correctness reference-answer structure. Second, for each of the three models, the mean CAP score by class largely follows the expected ordering among the relevant classes. (see \cref{tab:llm-semantic-distribution}).

\begin{table}[!tbh]
\centering
\scriptsize

\begin{tabular}{lccc}

\toprule

\rowcolor{diffbg}

\textbf{Class} & \textbf{GPT-4o} & \textbf{Gemini Flash} & \textbf{Qwen 3} \\

\midrule

exact                 & 111 & 84  & 48  \\

equivalent            & 268 & 232 & 296 \\

alternative-correct   & 235 & 102 & 242 \\

overinclusive-valid   & 213 & 114 & 225 \\

partial               & 37  & 251 & 73  \\

overinclusive-invalid & 6   & 7   & 16  \\

invalid               & 123 & 206 & 91  \\

contradictory         & 6   & 3   & 8   \\

\bottomrule

\end{tabular}

\caption{Number of human-labeled LLM-generated answers assigned to each correctness class.}

\label{tab:llm-semantic-counts}

\end{table}

Across the 25 strictly ordered class pairs, CAP incurs 4 violations for GPT-4o, 6 for Gemini Flash, and 2 for Qwen3-8B-Instruct. The most consistent deviation is the low score assigned to \texttt{alternative-correct}, which falls below its intended top-tier position for all three models. Gemini Flash additionally exhibits the previously observed \texttt{overinclusive-valid}/\texttt{partial} inversion. 
Overall, the class-level ordering from \cref{sec:monotonicity} largely generalizes to model-generated answers, while preserving the main failure modes identified on CAP-Correctness.



\section{Discussion}
\label{sec:discussion}
\subsection{Alignment with Human Judgement}
The empirical claim we make for CAP is that its scores follow the semantic correctness ordering induced by our taxonomy. While the taxonomy ordering is a benchmark design choice, the per-example class labels in CAP-Correctness were validated by human annotators against that taxonomy (\cref{app:dataset}). The monotonicity, pairwise accuracy, and ranking measures of \cref{sec:experiments} therefore assess alignment with that benchmark target.

The ``ground truth'' against which we benchmark metrics is itself constructed and inherits both the design choices of our taxonomy and the subjectivity of human annotation. We view this not as a flaw but as the unavoidable structure of semantic evaluation: there is no taxonomy-free notion of how correct an open-ended answer is. What the framework provides is an explicit, inspectable target ordering, against which any candidate metric---CAP, BLEU, COMET, or future learned alternatives---can be benchmarked on equal footing. The goal of the present work is correspondingly two-fold: (i) to show that CAP follows this operational ordering, and (ii) to show that it does so more reliably than existing metrics. 
The error analysis
(\cref{app:error_analysis}) confirms that
the alignment holds under human labels and breaks down only on
\texttt{alternative-correct} and the \texttt{partial}/\texttt{overinclusive-valid} pair.


\subsection{CAP as a Standalone Classifier}
A natural next step is to use CAP not only as a scorer but also as the labeler. Because CAP produces a continuous score in $[0, 1]$ that empirically separates the taxonomy classes (\cref{fig:comet_cap_distribution}), one can define class boundaries directly on the score axis--for example, $[0, 0.125)$ for contradictory, $[0.125, 0.25)$ for invalid, and so on up to the equivalent regime near $1$. Calibrating these thresholds on CAP-Correctness would yield a label-free evaluator that, given a question, a gold answer, and a prediction, returns both a continuous score and a taxonomy class.

This protocol also generalizes beyond CAP. A natural evaluation recipe for any future semantic correctness metric would be: (i) re-use CAP-Correctness as the benchmark, (ii) re-use the taxonomy ordering as the monotonicity target, (iii) calibrate the metric's own thresholds on the same labeled data, and (iv) reorder ambigious pairs according to the evaluation needs. Under this protocol, CAP-Correctness becomes a shared substrate for comparing semantic correctness metrics, independent of CAP itself.
\section{Conclusion and Future Work}\label{sec:conclusion}

We introduce a semantic correctness taxonomy for \openqa{} that replaces binary correctness with eight ordered classes capturing qualitatively different answer relations, including equivalence, incompleteness, valid elaboration, hallucinated additions, and contradiction. Beyond assigning more informative labels, the taxonomy provides a common structure for evaluating metrics: its ordering induces a monotonicity criterion for testing whether scores decrease as answers become semantically less correct. Within this framework, we introduce CAP, an NLI-based scorer that roughly doubles the rank correlation of the strongest baseline tested against the taxonomy ordering.

CAP's class-mean geometry also serves as a diagnostic of LLM answer style---rather than reporting only aggregate correctness, the distribution of outputs across classes reveals  model's tendencies. With threshold calibration, CAP functions as a label-free classifier returning both a score and a taxonomy class. The released datasets are reusable in their own right: CAP-Correctness as a labeled corpus future metrics can train against, and CAP-Statements as a QA-to-statement reformulation resource.

Several directions remain open for future work. Splitting answers into smaller subcomponents and scoring entailment over each could resolve the
bidirectional-NLI artefact on the \texttt{overinclusive-valid} / \texttt{partial} axis; stronger NLI backbones with broader world knowledge could close the \texttt{alternative-correct} ceiling; and multilingual NLI checkpoints could port the framework off English. The taxonomy itself is not fixed either: as LLMs evolve and \openqa{} expands into new domains, the categories that meaningfully partition model behavior will shift. CAP-Correctness and the monotonicity protocol support adding, splitting, or merging classes accordingly, rather than treating the current eight as final.

\section*{Limitations}
CAP inherits the limitations of reference-based, whole-statement NLI evaluation. It can structurally invert the \texttt{partial}/\texttt{overinclusive-valid} distinction, since partial answers are often entailed by the gold answer, while verbose valid answers often entail the gold but are not entailed by it. It can also under-score \texttt{alternative-correct} answers when a valid prediction satisfies the question without entailing the single reference answer, making CAP dependent on the NLI model's world knowledge.

Our benchmark is currently limited to English educational QA datasets derived from multiple-choice sources, with synthetic candidate answers and labels generated by a closed LLM-assisted pipeline. Although we validate a subset with human annotators, only a small portion of CAP-Correctness is human-labeled and examples are singly annotated, so we do not estimate inter-annotator agreement. The statement-generation step is another bottleneck, especially for long-context inputs, where reformulation errors can propagate directly into CAP scores.

Finally, CAP is also more computationally expensive than lightweight metrics, since each score requires statement generation and two NLI passes. The taxonomy assigns one label per answer, while real outputs may combine several correctness dimensions, such as partial correctness and unsupported extra information.

\section*{Ethics and Broader Impact}
\paragraph{Copyright and Licensing.}
CAP-Correctness contains question–answer content derived from OpenBookQA, AI2 ARC, and MMLU. AI2 ARC is distributed under CC BY-SA 4.0, while MMLU is distributed under the MIT License; the OpenBookQA repository is distributed under Apache 2.0, although the licensing status of the dataset content itself is not explicitly documented in the current dataset card. We retain source attribution and distribute source-derived content subject to the applicable original license terms.
\paragraph{Ethics and Data Privacy.} The source material consists of general-knowledge and academic questions and contains no personal, sensitive, or personally identifiable information. No student records, user identities, or private data are present, and the annotations are limited to educational QA content, posing no privacy risk.
\paragraph{Human Annotation.} Validation labels were collected from human annotators under the conditions described in \cref{app:annotation}.

\section*{Acknowledgments}

This work is partially supported by the project UNITe BG16RFPR002-1.014-0004 funded by PRIDST,  
and by the EU NextGenerationEU project, through the National Recovery and Resilience Plan of the Republic of Bulgaria, project SUMMIT, No.~BG-RRP-2.004-0008.

\bibliography{custom}

\clearpage
\appendix
\section{Boundedness of CAP}\label{app:cap-bounded}

We show that $\capfunction{\astatement}{\bstatement} \in [0,1]$ for
any pair of statements $(\astatement, \bstatement)$.

Since NLI probabilities sum to one for any ordered statement pair,
\begin{align}
    &P_{\text{entailment}}(\astatement \rightarrow \bstatement) +  \nonumber P_{\text{neutral}}(\astatement \rightarrow \bstatement) \\
    &+ P_{\text{contradiction}}(\astatement \rightarrow \bstatement) = 1.
\end{align}
and each probability lies in $[0,1]$. Subtracting the contradiction
term, which itself lies in $[0,1]$, 
\begin{align}
   0 \leq  P_{\text{contradiction}}(\astatement \rightarrow \bstatement) \leq 1.
\end{align}
as $\lambda \in [0,1]$, this yields
\begin{align}
    0 \;\leq\; & P_{\text{entailment}}(\astatement \rightarrow \bstatement)
    + \lambda P_{\text{neutral}}(\astatement \rightarrow \bstatement) \nonumber \\
    = \;& \jscore{\astatement}{\bstatement} \;\leq\; 1.
\end{align}
Because CAP is a convex combination of two directional scores
$\jscore{\astatement}{\bstatement}, \jscore{\bstatement}{\astatement}
\in [0,1]$ with weight $\alpha \in [0,1]$,
\begin{align}
    0 \;\leq\; \capfunction{\astatement}{\bstatement} \;\leq\; 1.
\end{align}
CAP attains its maximum when both directional entailment scores
approach $1$, corresponding to semantically equivalent statements,
and approaches $0$ when the NLI model assigns high contradiction
probability in both directions.

\section{Datasets details}\label{app:dataset}

\subsection{Annotation}\label{app:annotation}

Human annotation is used at three points in this work: validating a
1638-example subset of CAP-Correctness against the eight-class taxonomy
(\cref{app:cap-correctness}), validating $1{,}353$ generated declarative
statements from CAP-Statements (\cref{app:cap-statements}), and labeling
the LLM-generated answers used in \cref{sec:llm-eval}. All three
annotation tasks were performed by the same two annotators.

\paragraph{Annotator profile.} Both annotators hold bachelor's degrees
and certified C1-level English proficiency, and are therefore qualified
to judge the educational-level QA content used in our benchmarks. 

\paragraph{Compensation and consent. }Annotators were compensated at 3 times the average pay according to their demographic region. Prior to annotation, they were informed of the purpose of
the task, the intended research use of their labels, and the public
release of the resulting datasets and labels, and provided consent on
these terms. The annotated content consists exclusively of educational
questions and candidate answers and contains no sensitive, offensive,
or distressing material.

\subsection{Statement Generation Dataset}

\paragraph{Instructions.} Annotators received written guidelines
containing the relevant taxonomy (\cref{tab:semantic_labels} for the
correctness tasks, \cref{tab:statement-generation-labels} for statement
validation), one worked example per class, and a short calibration
batch before live annotation began.

\paragraph{Assignment.} Each example is assigned a single label by one
annotator. We do not double-annotate and therefore do not separately
estimate inter-annotator agreement; the agreement figures reported in
\cref{tab:agreement} measure human--LLM agreement, which is the
quantity of interest for validating the synthetic labeling pipeline.

\paragraph{Effort.} Across the three tasks, each annotator spent
approximately $12$ hours on annotation.

\subsection{CAP-Correctness}\label{app:cap-correctness} 
\begin{figure}[H]
    \centering
    \includegraphics[width=\columnwidth]{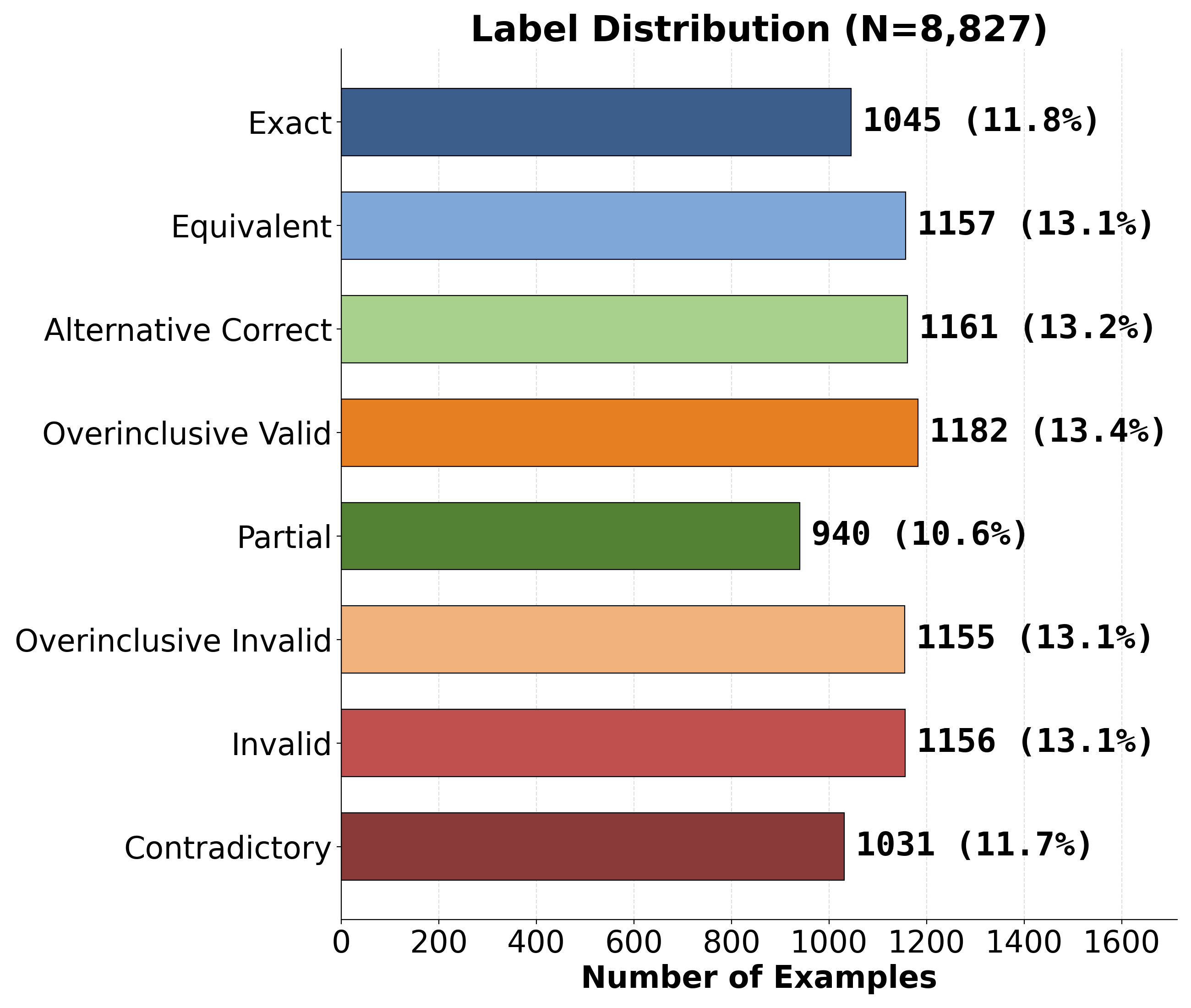}
    \caption{Distribution of semantic correctness labels in CAP-Correctness.}
    \label{fig:label-semantic}
\end{figure}

\paragraph{Generation Pipeline} For each \emph{[question, gold\_answer]} pair from the source corpora, the generation pipeline produces a single candidate answer. Generation is controlled to yield an approximately balanced distribution over the semantic labels describing the relationship between the gold answer and the predicted answer. Each label is instantiated with a separate prompt, conditioned on both the class definition in \cref{tab:semantic_labels} and the original gold answer. As a result, cases such as \texttt{overinclusive-valid} and \texttt{overinclusive-invalid} are generated through explicit, class-specific instructions rather than left to model discretion. We use the Claude Haiku 4.5 API. The prompt template and per-class
instructions are given in \cref{app:gen-prompt}. This class-conditioned setup explains the near-uniform per-class counts in \cref{fig:label-semantic}, in contrast to the long-tailed distribution expected from sampling natural model outputs.

\begin{figure}[H]
    \centering
    \small
    \includegraphics[width=\columnwidth]{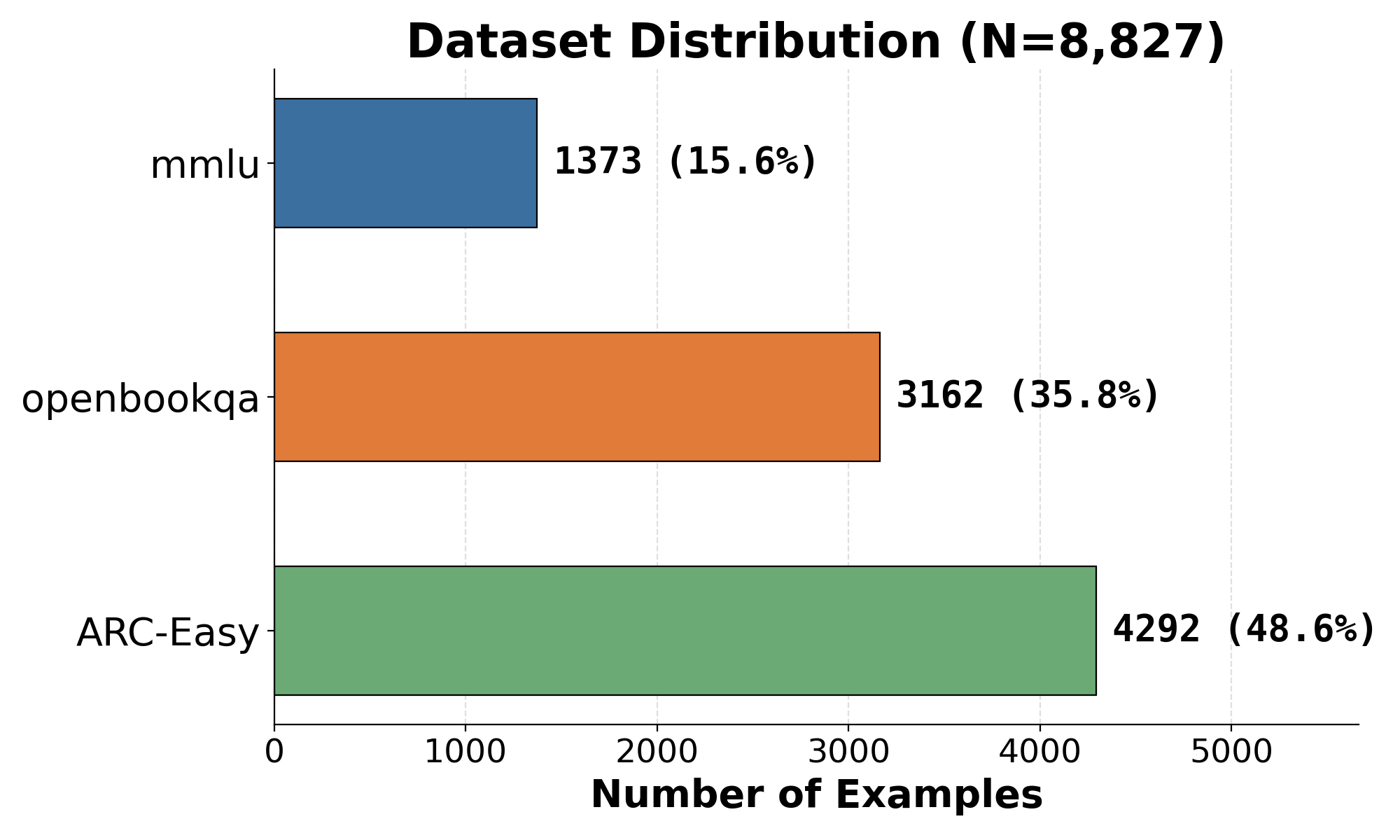}
    \caption{Questions acquisition statistics}
    \label{fig:dataset-acq-distr}
\end{figure}

\paragraph{Distribution.} \cref{tab:semantic_labels} lists the full set of correctness labels with their definitions, and \cref{fig:label-semantic} reports the resulting per-class counts on the 8{,}827 examples. The distribution is approximately balanced across all eight categories, with no class accounting for more than $\sim$13\% of the corpus, so downstream metric comparisons are not dominated by any single class. Acquisition per dataset is reported in \cref{fig:dataset-acq-distr}.

\begin{table}[h]
\centering
\setlength{\tabcolsep}{3pt}
\renewcommand{\arraystretch}{1.05}
\footnotesize
\begin{tabularx}{\columnwidth}{lX}
\toprule
\rowcolor{diffbg}
\textbf{Label} & \textbf{Definition} \\
\midrule

exact & Semantically identical to the gold answer. \\

equivalent & Same meaning expressed through linguistic variation or paraphrasing. \\

alternative-correct & Different but still semantically correct answer. \\

partial & Contains only part of the required information. \\

overinclusive-valid & Correct answer with additional valid information. \\

overinclusive-invalid & Correct answer with additional incorrect information. \\

invalid & Semantically incorrect answer. \\

contradictory & Explicitly contradicts the gold answer or question premise. \\

\bottomrule
\end{tabularx}
\caption{Semantic correctness labels used in the CAP evaluation benchmark.}
\label{tab:semantic_labels}
\end{table}

\paragraph{Validation Protocol} The 1638-example human validation uses the same eight-class taxonomy as the LLM-assisted pipeline (\cref{tab:semantic_labels}); annotation conditions and protocol are described in \cref{app:annotation}. Because each example receives a single human label, the figures in \cref{tab:agreement} quantify human--LLM agreement rather than inter-annotator agreement, which is appropriate for validating the synthetic labeling pipeline.

\paragraph{Interpretation.} 
The human--synthetic agreement is substantial under unweighted Cohen's $\kappa$ ($\kappa=0.779$) and increases under linear ($\kappa=0.825$) and quadratic ($\kappa=0.879$) weighting. Under the Landis--Koch convention \cite{landis1977measurement}, the weighted estimates fall in the almost-perfect agreement range. The increase under distance-sensitive weighting suggests that disagreements are disproportionately concentrated among categories that are closer in the taxonomy, rather than reflecting large shifts across the correctness ordering. Thus, even when the synthetic and human labels differ exactly, they often remain relatively close in terms of the semantic distinction encoded by the taxonomy. We take this as evidence that the synthetic labeling pipeline broadly aligns with human judgments under the proposed label scheme. This agreement, however, evaluates the reliability of the synthetic annotation pipeline given the taxonomy; it does not establish that the taxonomy itself is the correct model of semantic correctness, a separate question we return to in \cref{sec:discussion}.

\begin{table}[!tbh]
\centering
\small
\begin{tabular}{lc}
\toprule
\rowcolor{diffbg}
\textbf{Agreement Metric} & \textbf{Score} \\
\midrule
Cohen's $\kappa$ & 77.9 \\
Linear weighted $\kappa$ & 82.5 \\
Quadratic weighted $\kappa$ & 87.9 \\
\bottomrule
\end{tabular}
\caption{Human--LLM agreement on a 1638-example validation subset of the semantic correctness benchmark.}
\label{tab:agreement}
\end{table}

\subsection{Candidate-Answer Generation Prompt}\label{app:gen-prompt}

\paragraph{Design.} The candidate-answer generator uses a single
prompt template instantiated once per target class. Each call
receives the question, the gold answer, the target class name, the
class definition from \cref{tab:semantic_labels}, and one
class-specific instruction from \cref{tab:class-specific-instructions}.
Two design choices follow from the structure of the taxonomy. First,
conditioning generation on the gold answer (rather than asking the
model to produce its own gold) is what allows the overinclusive
classes to be split cleanly into the valid- and invalid-extras
variants: the model is told both what to preserve and what kind of
extra content to add. Second, the class-specific instruction is the
only component that varies across the eight calls per item, which
keeps the input/output schema and the surface-form constraints
constant across classes and avoids confounding the class signal with
formatting drift.

\paragraph{Prompt template.} The template below is sent for every
$(\question, \gold, \labelclass)$ triple; bracketed placeholders are
filled per call.
\begin{quote}\small\ttfamily\setlength{\parindent}{0pt}\raggedright
SYSTEM:\\
 f"Label: {label}"
        f"Definition: {label\_descriptions[label]}"
        "For each item below, generate one answer matching the label definition above."
        "Return ONLY a valid JSON array of strings, one answer per item, in the same order."
        "No explanation, no extra text — just the JSON array."
        f"Items:{json.dumps(items, indent=2)}"
        "JSON array:"
\end{quote}

\subsection{CAP-Statements}\label{app:cap-statements}
\subsubsection{Generation Pipeline}\label{app:statements-gen}
        
\begin{figure}[!tbh]
    \centering
    \includegraphics[width=\columnwidth]{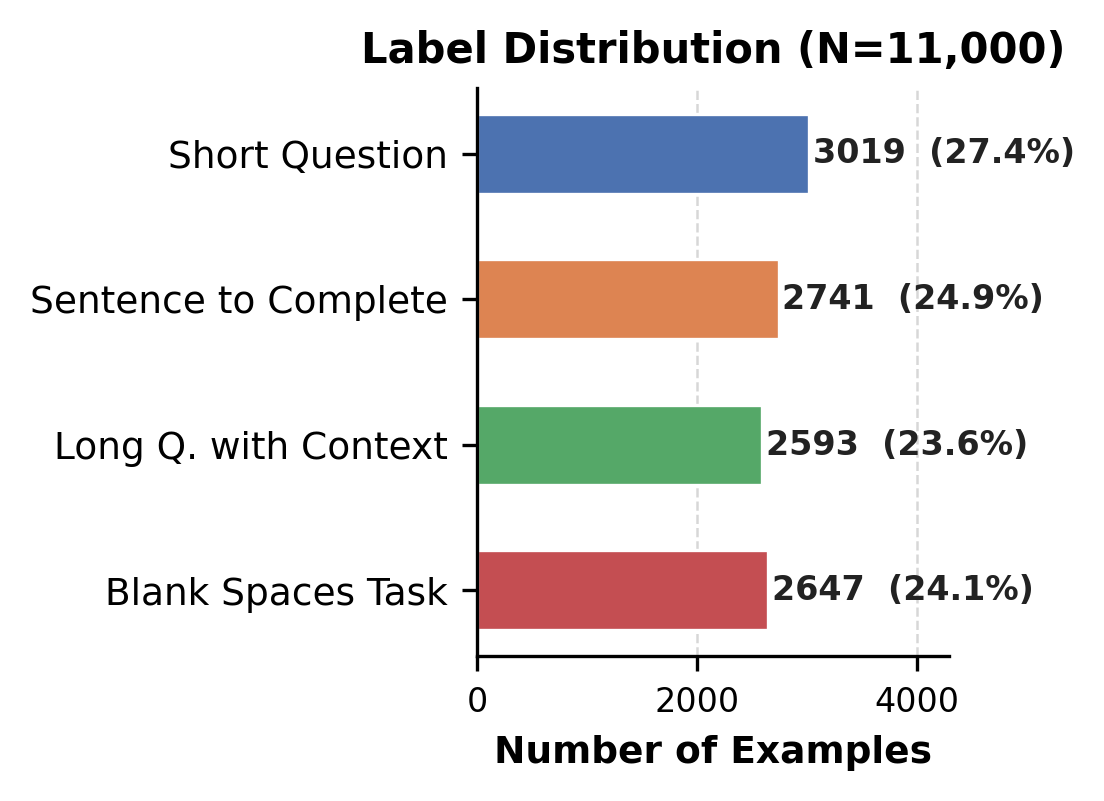}
    \caption{Distribution of question-type labels in CAP-Statements.}
    \label{fig:statements-distribution}
\end{figure}

\paragraph{Source-free synthesis.} Unlike CAP-Correctness, which
augments existing benchmark questions with candidate answers,
CAP-Statements is generated from scratch. We prompt Claude Sonnet 4.6 to produce
\emph{[question, answer, statement]} triples directly, conditioning
each call on (i) a target academic subject (general knowledge plus
seven academic subjects) and (ii) a target question type from
\cref{tab:question-type-labels}. This two-axis conditioning is what
yields the balanced question-type distribution reported
in \cref{fig:statements-distribution}, while ensuring controlled coverage across both dimensions.

\paragraph{Statement as supervision target.} The generator is
instructed to produce the declarative statement \emph{jointly} with
the question and answer, rather than reformulating an existing
question--answer pair post hoc. This guarantees that every training
example contains a statement that faithfully encodes the same
proposition expressed by the question--answer pair, removing one
source of label noise from the supervision used to fine-tune the
mT5-small reformulator.

\paragraph{Sampling and deduplication.} We sample with temperature
$1.0$ and \texttt{top\_p} $= 0.95$ to encourage lexical diversity,
issuing one API call per \emph{(subject, question-type)} cell and
generating in batches until the target cell count is reached.
Exact duplicates and near-duplicates (normalized-text Jaccard
$\geq 0.9$ on the question field) are removed. The resulting
11{,}000 triples are split into 8{,}800 / 1{,}100 / 1{,}100 for
train / validation / test (\cref{tab:datasetsshort}).

\paragraph{Prompt template.} The template below is sent for every
\emph{(subject, question-type)} cell; bracketed placeholders are
filled per call.

\begin{quote}\small\ttfamily\setlength{\parindent}{0pt}\raggedright
SYSTEM:\\
You generate training examples for a question-to-statement
reformulation model. Each example is a JSON object with exactly
three fields: \texttt{question}, \texttt{answer}, and
\texttt{statement}.\\[4pt]
The \texttt{statement} must be a single declarative sentence that
expresses the same proposition as the question--answer pair, with
no question marks, no second-person address, and no meta-commentary
(e.g., do not write ``the answer is\ldots'').\\[4pt]
Output ONLY the JSON object, with no preamble or surrounding text.\\[4pt]

USER:\\
Subject: \{subject\}\\
Question type: \{question\_type\}\\
Question-type definition: \{question\_type\_definition\}\\[4pt]
Generate one \emph{(question, answer, statement)} triple in the
specified subject and following the specified question type.
Constraints:\\
- The \texttt{question} must match the surface form of the
specified type (see definition above).\\
- The \texttt{answer} should be a phrase or short sentence; do not
explain or justify it.\\
- The \texttt{statement} must be a single grammatical declarative
sentence and must contain all the semantic content of the
question--answer pair, with no extra information.\\
- For the \texttt{blank-spaces-task} type, the \texttt{statement}
must fill in the blank explicitly (no underscores remain).\\
- For the \texttt{long-question} type, the \texttt{statement} must
compress all contextual sentences and the question into one
declarative clause.\\
- Do not produce duplicates of well-known textbook examples
(e.g., ``What gas do plants release during photosynthesis?''
already exists in the dataset).
\end{quote}

The \texttt{\{question\_type\_definition\}} placeholder is filled
from \cref{tab:question-type-labels}; the subject placeholder is
drawn uniformly from
\{\textit{general knowledge}, Biology, Physics, History, Geography,
Chemistry, Mathematics, Computer Science\} with a target proportion
of $56.2$\% general knowledge and $\approx 6.3$\% each for the
academic subjects.

The four question types are listed with definitions in \cref{tab:question-type-labels} and with concrete instances in \cref{tab:question-type-labels-examples}. The \texttt{short-question} type covers direct factoid questions, \texttt{sentence-to-complete} covers completion-style prompts, \texttt{long-question} covers items where the question is preceded by one or more sentences of context, and \texttt{blank-spaces-task} covers cloze-style items with explicit blanks. Long-question items are the most demanding regime for reformulation, since multiple sentences of context must be compressed into a single declarative claim while preserving the question--answer relation.

\begin{table}[!tbh]
\centering
\setlength{\tabcolsep}{3pt}
\renewcommand{\arraystretch}{1.05}
\footnotesize
\begin{tabularx}{\columnwidth}{lX}
\toprule
\rowcolor{diffbg}
\textbf{Label} & \textbf{Definition} \\
\midrule

short-question & Direct factual question. \\

sentence-to-complete & Sentence completion or prompt-style question. \\

long-question & Question preceded by contextual description or introductory sentences. \\

blank-spaces-task & Question or statement containing blank spaces that must be filled. \\

\bottomrule
\end{tabularx}
\caption{Label question type used in CAP-Statements}
\label{tab:question-type-labels}
\end{table}

\begin{table}[!tbh]
\centering
\setlength{\tabcolsep}{3pt}
\renewcommand{\arraystretch}{1.05}
\footnotesize
\begin{tabularx}{\columnwidth}{lX}
\toprule
\rowcolor{diffbg}
\textbf{Label} & \textbf{Example} \\
\midrule

short-question & What gas do plants release during photosynthesis? \\

sentence-to-complete & The sun is responsible for... \\

long-question & I have a shirt that is now too small, what can I do to conserve and reuse the fabric? \\

blank-spaces-task & Light bends when it passes from air into  \_\_\_\_\_\_ at an angle. \\
\bottomrule
\end{tabularx}
\caption{Examples for each question type used in CAP-Statements.}
\label{tab:question-type-labels-examples}
\end{table}

\paragraph{Distribution} \cref{fig:statements-distribution} reports the question-type distribution across the 11{,}000 triples. The four types are approximately balanced--no type accounts for more than 27.4\% of the corpus---ensuring that the statement generator is exposed to all reformulation regimes during training rather than being biased toward the easier short-question setting.

\paragraph{Validation Protocol.} To evaluate the quality of the resulting statement reformulations, we manually annotate a randomly sampled subset of 1{,}353 generated declarative statements (12.3\% of CAP-Statements); annotation conditions and protocol are described in \cref{app:annotation}. Each generated statement is assigned one of four labels--\textit{correct}, \textit{syntactic}, \textit{semantic}, or \textit{wrong}--following the taxonomy in \cref{tab:statement-generation-labels}.

\begin{table}[!tbh]
\centering
\small
\setlength{\tabcolsep}{4pt}
\renewcommand{\arraystretch}{1.05}
\begin{tabularx}{\columnwidth}{lX}
\toprule
\rowcolor{diffbg}
\textbf{Label} & \textbf{Definition} \\
\midrule
correct &
The generated statement preserves the full semantic meaning of the original question--answer pair without introducing grammatical or semantic errors.
\\

syntactic &
The generated statement contains syntactic or grammatical construction errors, but the underlying semantic meaning remains preserved.
\\

semantic &
The generated statement changes or distorts the semantic meaning of the original question--answer pair.
\\

wrong &
The generated statement fails as a faithful declarative reformulation, due to severe grammatical errors, semantic corruption, or both.
\\

\bottomrule
\end{tabularx}
\caption{Human evaluation taxonomy for generated declarative statements.}
\label{tab:statement-generation-labels}
\end{table}

\begin{figure}[!tbh]
    \centering
    \small
    \includegraphics[width=\columnwidth]{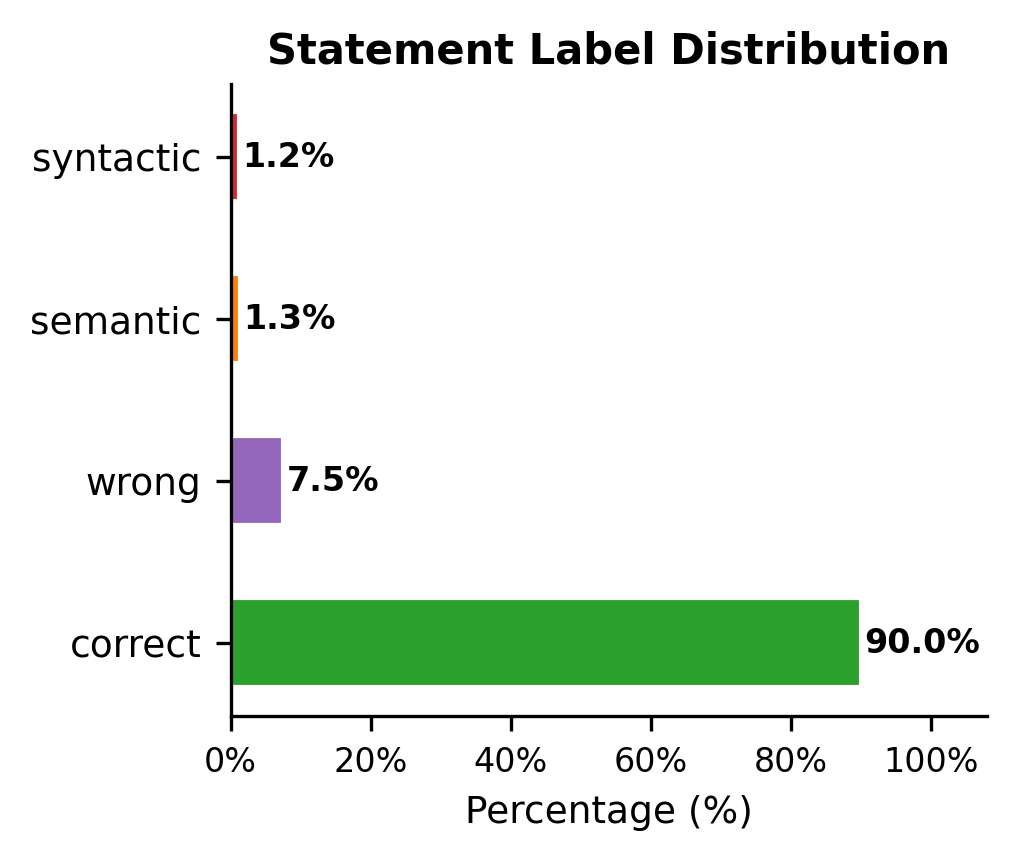}
    \caption{Errors in statements generation}
    \label{fig:distribution-statements-manual}
\end{figure}

Aggregated across question types (\cref{fig:distribution-statements-manual}), the manual labels indicate that the statement generation pipeline is generally reliable: 90.0\% of generated statements are labeled \textit{correct}, with isolated \textit{semantic} (1.3\%) and \textit{syntactic} (1.2\%) errors both rare. The largest error category is \textit{wrong} at 7.5\%, which by inspection consists overwhelmingly of cases where the model returns a near-verbatim concatenation of the question and the answer rather than a properly reformulated statement; the residual syntactic errors are dominated by omitted determiners (most often \emph{the}) and minor agreement mistakes that do not alter the underlying meaning. These patterns suggest the dominant failure mode is undertraining on harder reformulation cases rather than systematic semantic distortion. 

This aggregate picture obscures one important asymmetry across question types: long-question items are substantially harder for the reformulator than the other three. As reported in \cref{tab:question-type-impact}, 32.9\% of long-question statements are labeled \textit{wrong} and 9.6\% are labeled \textit{semantic}, compared to a combined error rate below 1\% on \texttt{sentence-to-complete} and \texttt{blank-spaces-task}. The practical consequence is that CAP's downstream reliability is question-type dependent: it is well-supported by the statement generator on short-question, sentence-completion, and blank-spaces items, and degraded on long-context items, where statement-reformulation errors propagate into the NLI scoring stage. We treat this as a limitation of the current statement generator rather than of the CAP framework itself, and discuss in \cref{sec:conclusion} natural directions for closing the gap--scaling the long-question training partition, or replacing mT5-small with a larger sequence-to-sequence reformulator.

\section{Experiment Details}\label{app:experiment-details}
\subsection{Evaluation Measure Definitions}
\label{app:eval-measures}
\paragraph{Rank correlation.} We compute Spearman's $\rho$ and
Kendall's $\tau$ between metric scores $\metric{\gold}{\pred}$ and the integer
ordinal rank $\rankfunc{\labelclass}$, treating tied taxonomy ranks (e.g.,
\texttt{exact} $\approx$ \texttt{equivalent} $\approx$
\texttt{alternative-correct}) using the standard mid-rank
convention.

\paragraph{Pairwise ranking accuracy.} Following the random baseline
of $0.5$, pairwise accuracy is defined as 
\begin{equation}
\pairacc{\metricempty} = \mathbb{E}_{(a, b)\sim \mathcal{D}}
\bigl[ \indicator{\metricempty(a) > \metricempty(b)} \bigr],
\end{equation}
i.e., the fraction of class-ordered pairs on which the metric assigns a higher score to the more correct answer. Pairs $(a, b)$ are sampled such that $\rankfunc{\labelclass_{a}} > \rankfunc{\labelclass_{b}}$ under the strict (non-tied) part of the taxonomy ordering.

\paragraph{Monotonicity violations.} Let $\meanscore{\metricempty}{\labelclass}
$ denote the mean
metric score on class $\labelclass$. The taxonomy induces 25 strictly ordered
class pairs. A pair $(\labelclass_i, \labelclass_j)$ with $\labelclass_i \succ \labelclass_j$ is a
\emph{violation} if $\meanscore{\metricempty}{\labelclass_i} \leq \meanscore{\metricempty}{\labelclass_j}$. The violation rate
$\violationrate{\metricempty} \in [0, 1]$ is the fraction of violating pairs out of $25$.

\paragraph{Hard neighboring-pair accuracy.} To isolate locally
challenging distinctions, we additionally report pairwise accuracy
restricted to adjacent classes in \cref{eq:correctness-order}, focusing
on the four main contrasts reported in Tab.~\ref{tab:hard-pairwise-ranking}.

\subsection{Implementation Details}\label{app:implementation-details}
\paragraph{NLI backbone.} CAP scores are computed with \texttt{cross-encoder/nli-deberta-v3-large}, a publicly available NLI cross-encoder from the Hugging Face Hub. We additionally evaluated \texttt{MoritzLaurer/mDeBERTa-v3-base-mnli-xnli} and \texttt{joeddav/xlm-roberta-large-xnli} during development but selected \texttt{nli-deberta-v3-large} for its sharper contradiction-class separation on the CAP-Correctness validation subset. Inputs are tokenized as $(\text{premise}, \text{hypothesis})$ pairs with a maximum joint length of $512$ tokens. Entailment, neutral, and contradiction probabilities are read from the softmax output of the classification head. Inference is run in mixed precision (\texttt{torch.float16}) with batch size $16$.

\paragraph{Directional scoring.} For each $(\question, \gold, \pred)$ triple, CAP requires two NLI passes: forward ($\gold \rightarrow \pred$) with $\statement{\question}{\gold}$ as premise and $\statement{\question}{\pred}$ as hypothesis, and reverse ($\pred \rightarrow \gold$) with the two swapped. Both passes use the same NLI checkpoint without further fine-tuning.

\paragraph{CAP hyperparameters.} Unless stated otherwise we use $\alpha = 0.85$, $\lambda = 0.3$, selected on a held-out subset of CAP-Correctness; the full sweep is reported in \cref{tab:cap-hyperparameters}.

\paragraph{Ordinal rank.} The taxonomy ordering of \cref{eq:correctness-order} is realized as an integer severity map when computing rank correlations and monotonicity violations: \texttt{exact} $= 6$, \texttt{equivalent} = \texttt{alternative-correct} $= 6$, \texttt{overinclusive-valid} $= 5$, \texttt{partial} $= 4$, \texttt{overinclusive-invalid} $= 2$, \texttt{invalid} $= 1$, \texttt{contradictory} $= 0$. Tied severities (\texttt{equivalent} vs.\ \texttt{alternative-correct}) contribute no strict pair and are not counted toward the monotonicity total. Spearman and Kendall correlations use the standard mid-rank convention for tied ordinal ranks.


\paragraph{Statement generator.} We fine-tune mT5-small \citep{xue-etal-2021-mt5} on the $8{,}800$ training examples of CAP-Statements with the AdamW optimizer, learning rate $5\!\times\!10^{-5}$, batch size $4$ with accumulation $2$, and a maximum of $15$ epochs with early stopping on validation ROUGE-L. Decoding uses beam search with width $4$ and a maximum output length of $512$ tokens. Final test performance is reported in \cref{sec:datasets}.

\paragraph{Baseline metric versions.} BLEU and METEOR are computed with \texttt{nltk.translate} (\texttt{sentence\_bleu} and \texttt{meteor\_score} respectively); ROUGE-L is computed with the
\texttt{rouge\_score} package (\texttt{rouge\_scorer}, \texttt{rougeL}
variant). BERTScore uses the F1 variant via the \texttt{bert\_score}
package with its default English checkpoint. COMET scores are
computed via the \texttt{comet} package with the
\texttt{Unbabel/wmt22-comet-da} checkpoint.

\paragraph{LLM-generated answers.} For \cref{sec:llm-eval} we collect free-form answers to a 1k sample from CAP-Correctness questions from GPT-4o (\texttt{gpt-4o-2024-08-06}), Gemini 2.0 Flash (\texttt{gemini-2.0-flash}), and Qwen3-8B-Instruct (\texttt{Qwen/Qwen3-8B-Instruct}), each prompted in a zero-shot setting with temperature $0.0$ and a $256$-token output cap.

\paragraph{Prompt template.} The same template is issued to every
model and every question; the \texttt{\{question\}} placeholder is the
only field that varies per call.

\begin{quote}\small\ttfamily\setlength{\parindent}{0pt}\raggedright
SYSTEM:\\
Answer the question in the style of a person responding on an exam:
concise, factually correct, and free of elaboration. Use a single
short sentence when sufficient; never exceed three sentences. Do not
restate the question, include preamble or qualifications, or open with
phrases such as ``the answer is\ldots''. Respond in plain prose
without any markdown formatting: no bold, italics, headers, lists, or
code blocks.\\[4pt]

USER:\\
\{question\}
\end{quote}

\paragraph{Hardware.} All NLI inference and mT5 fine-tuning are run on a single NVIDIA A100 40\,GB GPU; no experiment requires distributed training.

\subsection{CAP Hyperparameter Sweep}\label{app:cap-sweep}

We select $\alpha$ and $\lambda$ via a grid sweep on a 1000 held-out
subset of CAP-Correctness, evaluating each setting on Spearman $\rho$,
pairwise ranking accuracy, and monotonicity violation count
(\cref{tab:cap-hyperparameters}). Moderate neutral weighting
($\lambda = 0.3$) combined with asymmetric scoring ($\alpha = 0.85$)
achieves the best ranking and pairwise accuracy while matching the
lowest violation count, outperforming both purely entailment-based
($\lambda = 0$) and purely unidirectional ($\alpha = 1$) variants.
This supports the intuition that partial credit for neutral relations
improves robustness on semantically incomplete and alternative-correct
answers while preserving strong contradiction separation.

\begin{table}[H]
\centering
\small
\setlength{\tabcolsep}{4pt}
\renewcommand{\arraystretch}{1.05}
\begin{tabular}{ccccc}
\toprule
\rowcolor{diffbg}
$\alpha$ & $\lambda$ & Spearman $\rho$ & PairAcc & Violations \\
\midrule
0.3 & 0.0 & 54.20 & 74.16 & 4/25 \\
0.5 & 0.0 & 56.36 & 75.33 & 4/25 \\
0.5 & 0.3 & 57.96 & 76.20 & 4/25 \\
0.7 & 0.3 & 59.33 & 76.92 & 4/25 \\
\textbf{0.85} & \textbf{0.3} & \textbf{59.41} & \textbf{76.97} & \textbf{4/25} \\
\bottomrule
\end{tabular}
\caption{Effect of directional asymmetry ($\alpha$) and neutral-class weighting ($\lambda$) on semantic ranking consistency.}
\label{tab:cap-hyperparameters}
\end{table}
\section{Results}
\subsection{Confidence intervals}\label{app:confidence}

The relatively narrow confidence intervals indicate stable correlation estimates; moreover, CAP’s intervals ($[58.56, 62.11]$ for Spearman and $[47.32, 50.35]$ for Kendall) remain well separated from those of all baselines, supporting the robustness of its ranking advantage (\cref{tab:ranking-bootstrap-ci}).

\begin{table}[!tbh]
\centering
\setlength{\tabcolsep}{3pt}
\renewcommand{\arraystretch}{1.05}
\scriptsize

\begin{tabular}{@{}lcccc@{}}
\toprule
\rowcolor{diffbg}
\textbf{Metric}
& \multicolumn{2}{c}{\textbf{Spearman $\rho \times 100$}}
& \multicolumn{2}{c}{\textbf{Kendall $\tau \times 100$}} \\
\cmidrule(lr){2-3}\cmidrule(lr){4-5}

\rowcolor{diffbg}
& \textbf{Score}
& \textbf{95\% CI}
& \textbf{Score}
& \textbf{95\% CI} \\
\midrule

BLEU
& 10.95 & [8.55, 13.25]
& 8.31 & [6.41, 10.12] \\

ROUGE-L
& 16.67 & [14.29, 19.08]
& 13.00 & [11.04, 14.95] \\

METEOR
& 14.70 & [12.41, 17.01]
& 11.45 & [9.62, 13.27] \\

BERTScore
& 16.10 & [13.82, 18.39]
& 10.90 & [9.22, 12.63] \\

COMET
& 26.88 & [24.77, 28.99]
& 19.57 & [17.98, 21.18] \\

\midrule
\rowcolor{greenbg}
\textbf{CAP}
& \textbf{60.37} & \textbf{[58.56, 62.11]}
& \textbf{48.83} & \textbf{[47.32, 50.35]} \\

\bottomrule
\end{tabular}

\caption{
We report 95\% bootstrap confidence intervals computed using 10,000 resamples from the 7,827 test examples in CAP-Correctness.
}
\label{tab:ranking-bootstrap-ci}
\end{table}

\subsection{Monotonicity Analysis}\label{app:monotonicity}


While the main results show that CAP better captures the taxonomy’s global ranking structure, monotonicity asks a sharper question: does CAP preserve the taxonomy’s ordering on a class-by-class basis? Recall from \cref{sec:eval-measures} that a metric incurs a violation on a strictly ordered pair $(\labelclass_i, \labelclass_j)$ whenever the mean score for the higher-severity class falls below the mean score for the lower-severity one. 
 

Lexical metrics and BERTScore invert roughly half of the ordering. Even COMET, the strongest baseline, violates $9/25$ comparisons it is expected to respect. CAP reduces this count to $4/25$: on $21$ of the $25$ ordered pairs, the mean CAP score for the more correct class is higher than for the less correct one (see \cref{tab:monotonicity-violations}).


\cref{fig:comet_cap_distribution} makes the score geometry visible. CAP’s per-class distributions occupy progressively lower ranges as we move down the taxonomy, with classes forming largely distinct bands on the score axis. COMET’s distributions, by contrast, collapse into a narrow region with heavy overlap across nearly every class: only \texttt{exact} separates cleanly, while \texttt{equivalent}, \texttt{partial}, the overinclusive classes, and even \texttt{invalid} answers occupy largely the same range. CAP’s monotone ordering is visible on the score axis; COMET does not exhibit a comparable ordering.

The four class-pair inversions made by CAP,
reflect two distinct failure modes. First, \texttt{alternative-correct} scores below the other top-tier classes. This class is difficult for NLI-based scoring because a valid alternative need not entail the gold answer, nor be entailed by it: for example, \textit{Jupiter} and \textit{Mars} may both be valid answers in context, while remaining mutually non-entailing. This effect is amplified by the pretrained NLI model’s uneven world knowledge across CAP-Correctness topics, which can prevent it from recognizing valid alternatives. The second failure mode is \texttt{partial} scores above their expected positions, a structural artifact of bidirectional NLI that we examine in detail in the next section. Despite these localized inversions, CAP preserves the taxonomy’s ordering far more consistently than any of the baselines.

\begin{table}[H]
\centering
\setlength{\tabcolsep}{5pt}
\renewcommand{\arraystretch}{1.05}
\footnotesize
\begin{tabular}{lcc}
\toprule
\rowcolor{diffbg}
\textbf{Metric} & \textbf{Violating Class Pairs} & \textbf{Violation Rate} \\
\midrule
BLEU      & 14 / 25 & 56.00 \\
ROUGE-L   & 12 / 25 & 48.00 \\
METEOR    & 12 / 25 & 48.00 \\
BERTScore & 11 / 25 & 44.00 \\
COMET     & 9  / 25 & 36.00 \\
\midrule
\rowcolor{greenbg}
\textbf{CAP} & \textbf{4 / 25} & \textbf{16.00} \\
\bottomrule
\end{tabular}
\caption{Monotonicity violations, defined as cases where the mean score of a lower-severity class exceeds the mean score of a higher-severity class.}
\label{tab:monotonicity-violations}
\end{table}
\section{Error Analysis} \label{app:error_analysis}
We dissect CAP's residual errors along three axes: statement-generation quality (\cref{tab:question-type-impact}), label-source stability (\cref{tab:label-source-stability,tab:per-class-label-source}), and the per-class structure of the failure modes identified in \cref{sec:results}.

\paragraph{Statement-generation reliability across question types}

CAP's reformulation step relies on the mT5 statement generator, whose error rate is markedly uneven across question types (\cref{tab:question-type-impact}; full breakdown in \cref{app:dataset}). Sentence-completion and blank-spaces items are produced cleanly, short-question items show a moderate \textit{wrong} rate; long-context items are the clear bottleneck at $32.9\%$ \textit{wrong} and $9.6\%$ \textit{semantic} errors. Downstream CAP scores therefore inherit a question-type-dependent reliability, with the dominant source of error attributable to the statement generator on long-context inputs rather than to the metric itself.

\begin{table}[!tbh]
\centering
\small
\begin{tabular}{lccc}
\toprule
\rowcolor{diffbg}
\textbf{Question Type} & \textbf{syntactic} & \textbf{semantic} & \textbf{wrong} \\
\midrule
Short question & 3.0\% & 0.6\% & 10.5\% \\
Long context & 3.6\% & 9.6\% & 32.9\% \\
Sentence completion & 0.2\% & 0.0\% & 0.0\% \\
Blank-space task & 0.4\% & 0.0\% & 0.0\% \\
\bottomrule
\end{tabular}
\caption{Human validation of generated statements by question type. Percentages indicate the proportion of statements labeled syntactic, semantic, or wrong during manual evaluation.}
\label{tab:question-type-impact}
\end{table}

\subsection*{Stability under synthetic vs.\ human labels}
We re-evaluate CAP on the 1,638-example human-validated subset under both synthetic and human-assigned labels. CAP remains stable across label sources (\cref{tab:label-source-stability}), with slightly higher performance under human labels ($\rho$: 69.89 vs.\ 66.87; $\tau$: 57.34 vs.\ 54.69; PairAcc: 87.16 vs.\ 84.87). This consistency indicates that CAP's ranking performance is not an artifact of the synthetic annotation pipeline; if anything, the synthetic labels provide a slightly more conservative estimate of its alignment with the human-labeled taxonomy ordering.


\begin{table}[!tbh]
\centering
\small
\begin{tabular}{lccc}
\toprule
\rowcolor{diffbg}
\textbf{Label Source} & \textbf{Spearman $\rho$} & \textbf{Kendall $\tau$} & \textbf{PairAcc} \\
\midrule
Synthetic & 66.87 & 54.69 & 84.87 \\
Human-labeled & 69.89 & 57.34 & 87.16 \\
\bottomrule
\end{tabular}
\caption{Stability of CAP evaluation under synthetic and human semantic labels.}
\label{tab:label-source-stability}
\end{table}


\subsection*{Label-semantic synthetically generated vs human label}
The per-class CAP means are highly consistent across synthetic and human label sources (\cref{tab:per-class-label-source}). For most classes, the difference is within a few score points, and both labelings recover the same overall score structure: \texttt{exact} and \texttt{equivalent} score highest, \texttt{invalid} and \texttt{contradictory} lowest, with the intermediate classes occupying similar positions. The largest shifts occur for \texttt{partial} (83.61 vs.\ 77.98) and \texttt{alternative-correct} (36.74 vs.\ 31.74), but these do not change the broader pattern. This suggests that CAP's class-level behavior is largely stable to whether the taxonomy labels are synthetic or human-assigned.
\begin{table}[!tbh]
\centering
\small
\begin{tabular}{lcc}
\toprule
\rowcolor{diffbg}
\textbf{Class} & \textbf{Synthetic} & \textbf{Annotator}  \\
\midrule
Exact & 98.79 & 98.62  \\
Equivalent & 84.59 & 86.76 \\
Partial & 83.61 & 77.98 \\
OV & 50.07 & 49.91 \\
OI & 39.20 & 38.46\\
Invalid & 13.04 & 14.56 \\
Contradictory & 3.18 & 4.22 \\
Alt. & 36.74 & 31.74\\
\bottomrule
\end{tabular}
\caption{Mean CAP scores per semantic class under synthetic and human label sources.}
\label{tab:per-class-label-source}
\end{table}

\subsection*{Per-class failure modes}

The per-class means in \cref{tab:per-class-label-source} concentrate CAP's residual error in two locations. \texttt{Partial} sits above its expected taxonomic position (mean $83.61$ synthetic, $77.98$ annotator), and \texttt{alternative-correct} sits below the top group (mean $36.74$ synthetic, $31.74$ annotator). The remaining classes--including \texttt{overinclusive-invalid}--occupy roughly their expected positions.

\paragraph{Partial scores too high.} The mechanism is the structural NLI artefact described in \cref{sec:hard-pairs}. Partial answers are subsets of the gold information, so the dominant $\gold \rightarrow \pred$ direction--carrying weight $\alpha = 0.85$--measures whether the gold entails the partial answer, which it largely does. The reverse direction $\pred \rightarrow \gold$ correctly fails to entail (the partial is missing content) but carries only $1-\alpha = 0.15$ of the weight. CAP therefore inherits the high gold-side entailment and ranks partial answers near the top group rather than mid-taxonomy.

\paragraph{Overinclusive-valid scores too low.} \texttt{Overinclusive-valid} is the mirror case. The answer contains the gold plus additional valid content, so $\pred \rightarrow \gold$ entails strongly--but this is the down-weighted direction. The dominant $\gold \rightarrow \pred$ direction sees gold-plus-extras and assigns mostly neutral mass, since the extras are not supported by the gold. The asymmetric weighting amplifies this neutral signal and pushes \texttt{overinclusive-valid} below \texttt{partial}, inverting the taxonomy on this specific axis.

\paragraph{Overinclusive-invalid is positioned correctly.} In contrast to the two cases above, \texttt{overinclusive-invalid} behaves as expected. Its mean ($39.20$ synthetic, $38.46$ annotator) sits below \texttt{overinclusive-valid} because the invalid extras introduce contradiction mass that the entailment + $\lambda \cdot$~neutral score does not reward, and clearly above \texttt{invalid} and \texttt{contradictory} because the correct content is still recognized. CAP separates ``correct with false extras'' from ``correct with valid extras'' and from ``fully invalid'' in the expected order. The two overinclusive classes are difficult for surface-similarity baselines (\cref{tab:hard-pairwise-ranking}, OV $>$ OI column), but they are not where CAP itself fails.



\begin{table}[!tbh]
\centering
\setlength{\tabcolsep}{3pt}
\renewcommand{\arraystretch}{1.1}
\footnotesize
\begin{tabularx}{\columnwidth}{lX}
\toprule
\rowcolor{diffbg}
\textbf{Class} & \textbf{Class-specific instruction} \\
\midrule
exact &
Reproduce the gold answer verbatim. \\
equivalent &
Express the same meaning as the gold answer using different
wording. Do not omit or add information. \\
alternative-correct &
Produce a different but equally valid answer to the question. The
answer must be factually correct and must not be entailed by the
gold answer in either direction. \\
partial &
Reproduce a strict subset of the information in the gold answer.
Do not add information beyond what the gold states. \\
overinclusive-valid &
Include the full content of the gold answer and add one short piece
of additional information that is factually correct and relevant
to the question. \\
overinclusive-invalid &
Include the full content of the gold answer and add one short piece
of additional information that is factually incorrect or
unsupported. \\
invalid &
Produce a plausible-sounding answer to the question that is
factually wrong. Do not directly contradict the gold answer. \\
contradictory &
Produce an answer that directly contradicts the gold answer or
rejects the premise of the question. \\
\bottomrule
\end{tabularx}
\caption{Class-specific instructions used to instantiate the
candidate-answer generation prompt. The \{class\_instruction\}
placeholder in the template is filled with the row matching the
target class.}
\label{tab:class-specific-instructions}
\end{table}

\end{document}